\documentclass{article}
\usepackage{ijcai25}
\usepackage{amsfonts}
\usepackage{times}
\usepackage{soul}
\usepackage{url}
\usepackage[hidelinks]{hyperref}
\usepackage[utf8]{inputenc}
\usepackage[small]{caption}
\usepackage{graphicx}
\usepackage{amsmath}
\usepackage{amsthm}
\usepackage{booktabs}
\usepackage{algorithm}
\usepackage{algorithmic}
\usepackage[switch]{lineno}
\usepackage{subcaption}
\usepackage{multirow}
\usepackage{lipsum}
\usepackage{xcolor}

\newcommand{\method}{\texttt{MaTU}}

\title{Many-Task Federated Fine-Tuning via Unified Task Vectors}

\author{
    Vasileios Tsouvalas$^1$\ \and Tanir Ozcelebi \And Nirvana Meratnia$^1$
    \affiliations $^1$Eindhoven University of Technology\\
    \emails \{v.tsouvalas, t.ozcelebi, n.meratnia\}@tue.nl
}

\begin{document}

\maketitle

\begin{abstract}
Federated Learning (FL) traditionally assumes homogeneous client tasks; however, in real-world scenarios, clients often specialize in diverse tasks, introducing \textit{task heterogeneity}. To address this challenge, Many-Task FL (MaT-FL) has emerged, enabling clients to collaborate effectively despite task diversity. Existing MaT-FL approaches rely on client grouping or personalized layers, requiring the server to manage individual models and failing to account for clients handling multiple tasks. We propose~\method, a MaT-FL approach that enables joint learning of task vectors across clients, eliminating the need for clustering or client-specific weight storage at the server.~\method~constructs a single ``\textit{unified}'' task vector per client and leverages lightweight task-specific modulators — binary masks and scalar rescalers — to disentangle and adapt per-task behavior. This design ensures stateless server aggregation and naturally scales to large task sets with minimal communication overhead. Evaluated across 30 datasets,~\method~achieves superior performance over state-of-the-art MaT-FL approaches, with results comparable to per-task fine-tuning, while delivering significant communication savings.
\end{abstract}


\vspace{-5pt}
\section{Introduction}\label{sec:introduction}

Federated Learning (FL) has emerged as a collaborative learning paradigm, enabling joint training of neural network models across edge devices (refereed to as clients), while keeping data localized~\cite{mcmahan2017communication}. Traditional FL settings assume clients work on the same set of tasks; yet, in real-world cross-silo FL the inherent task heterogeneity among clients can naturally occur as clients typically specialize in specific tasks. This phenomenon, known as \textit{task heterogeneity}, represents a novel and under-explored form of heterogeneity in FL, adding complexity to the learning process.

Recent studies~\cite{chen2023fedbone,zhuang2023mas,10208649} have underscored the significance of \textit{task heterogeneity} in FL. To address this diversity, traditional FL has evolved into Many-Task Federated Learning (MaT-FL)~\cite{10208649} to enable clients to effectively collaborate, despite specializing in different tasks. It is worth to mention that while ``\textit{many-task}'' has been interchangeably used as ``\textit{multi-task}'' in conventional ML, early Multi-Task Federated Learning (MTFL) mainly addressed personalized FL, a subset of MaT-FL. MaT-FL is thus used to distinguish it from prior MTFL works~\cite{10208649,muhamed2024fed,zhuang2023mas}. Current MaT-FL approaches address \textit{task heterogeneity} by focusing on client grouping or split FL. For instance, FedBone~\cite{chen2023fedbone} employs split FL whereas the server needs to sustain a unique model per client. Conversely, MaT-FL~\cite{10208649} and MAS~\cite{zhuang2023mas} concentrate on dynamically grouping client models based on task similarity, aggregating models among clients that handle similar tasks. Nevertheless, these strategies require the server to manage individual models for each client or task and fail to address scenarios where clients hold more than one task (
e.g., a personal health tracker monitoring multiple biomarkers such as heart-rate, blood-pressure, blood glucose level, body temperature simultaneously
). Training a single model among tasks offers considerable benefits in federated settings. Training individual models for each task is resource-intensive, whereas a unified model not only conserves resources but also potentially enhances performance by leveraging learning from diverse auxiliary tasks. 

Concurrently, the advent of large-scale deep models trained on massive amounts of data in a self-supervised fashion has ushered in a new era of Foundation Models (FMs). These models, exemplified by Large Language Models (LLMs) and vision Foundation Models (vFMs) hold the promise of adapting to a wide range of downstream tasks. Furthermore, Parameter Efficient Fine-Tuning (PEFT) techniques, allowing FMs to reach performance levels comparable with fully fine-tuned models with only minimal adjustments to their parameters, have recently seen immense popularity, with LoRA adapters standing out for their efficacy and broad applicability in FL~\cite{yi2023fedlora,cho2024heterogeneous,nguyen2024flora,ping2024fl}. 

This paper addresses the challenge of adapting pretrained Foundation Models (FMs) to the MaT-FL setting, where clients hold an arbitrary (\textgreater$1$) number of tasks, and we aim to train a unified model across all tasks. We propose \textbf{\underline{Ma}}ny-\textbf{\underline{T}}ask FL via \textbf{\underline{U}}nified Task Vectors (\method), a novel approach that eliminates the need for clustering or client-specific weight storage at the server, enabling natural knowledge transfer across tasks in FL. Inspired by recent advancements in model merging and Task Arithmetic (TA)~\cite{ilharco2023editing,yadav2023tiesmerging},~\method~introduces a task vector aggregation mechanism tailored for MaT-FL. Specifically, we determine task similarity directly based on the direction of clients' task vectors and constructs a ``\textit{unified}'' task vector that encapsulates all tasks. To handle task-specific requirements, we augment the ``\textit{unified}'' vector with lightweight modulators (i.e., binary task-specific masks and scalers) that facilitate knowledge transfer between related tasks while promoting weight disentanglement for dissimilar ones, ensuring robust performance across heterogeneous tasks in FL. By transmitting only a single vector and lightweight modulators per client,~\method~avoids the per-task overhead of existing MaT-FL methods, enabling both communication efficiency and scalability to large task sets. Concisely, our main contributions are as follows:

\begin{itemize}
\item We introduce~\method, a MaT-FL approach enabling joint training across clients which holding multiple tasks without requiring server-side clustering or client-specific weight storage.
\item We propose a novel task vector aggregation mechanism that constructs a unified task vector across tasks, while leveraging lightweight task-specific modulators to promote weight disentanglement and knowledge transfer.
\item We conduct a comprehensive performance evaluation across 30 datasets, where~\method~outperforms state-of-the-art MaT-FL regimes with performance comparable to per-task adaptation fine-tuning, all while delivering significant communication savings.
\end{itemize}

\section{Related Work}

\noindent \textbf{Federated PEFT.} Parameter Efficient Fine-Tuning (PEFT) has emerged as an alternative fine-tuning strategy, where most of the pre-trained model parameters are frozen, and only a small portion of task-specific parameters are trained. Among the various PEFT methods, the injection of additional adapter modules, fine-tuned independently from the pre-trained model — such as Low-Rank Adaptation (LoRA)\cite{hu2021loralowrankadaptationlarge} — has achieved state-of-the-art results across many large pre-trained models. 

In the context of FL, PEFT-based approaches have demonstrated success in enhancing privacy\cite{sun2024improvingloraprivacypreservingfederated}, communication~\cite{tsouvalas2023federatedfinetuningfoundationmodels}, and computation~\cite{babakniya2023slorafederatedparameterefficient} efficiency. SLoRA~\cite{babakniya2023slorafederatedparameterefficient} addresses data heterogeneity in federated settings by utilizing multiple LoRAs, while FFA-LoRA~\cite{sun2024improvingloraprivacypreservingfederated} enhances privacy in FL by applying differential privacy techniques to LoRA modules. Moreover, DeltaMask~\cite{tsouvalas2023federatedfinetuningfoundationmodels} combines LoRAs with probabilistic encoding to achieve significant bitrate reductions during FMs fine-tuning in FL, and FS-LLM~\cite{kuang2023federatedscopellmcomprehensivepackagefinetuning} extends PEFT to LLMs in the federated settings. Nonetheless, most PEFT-based FL approaches primarily focus on single-task settings, overlooking their potential in many-task settings. In our work, we explore combining LoRAs with model merging techniques, such as TIES~\cite{yadav2023tiesmergingresolvinginterferencemerging}, to address \textit{task heterogeneity} in FL, enabling the training of a unified model across multiple tasks while achieving significant reductions in communication and computation overhead. \\

\noindent \textbf{MaT-FL.} Many-Task Federated Learning (MaT-FL) has emerged to tackle \textit{task heterogeneity} in FL. From MTFL approaches, FedProx~\cite{li2020federatedoptimizationheterogeneousnetworks}, introduced a proximity term to limit client updates from deviating too far from the global model in MTFL. Alternatively, FedBone~\cite{chen2023fedbone} used split FL to introduce task-specific personalized layers to handle multiple tasks in FL. Recent MaT-FL approaches primarily address \textit{task heterogeneity} by focusing on client grouping~\cite{10208649,zhuang2023mas,lu2023towards}. MaT-FL~\cite{10208649} and MAS~\cite{zhuang2023mas} dynamically group clients based on task similarity, enabling model aggregation among clients with similar tasks. FedHCA~\cite{lu2023towards} extends this by supporting a variable number of tasks across clients, creating a more flexible framework. More recently, NTKFedAvg~\cite{muhamed2024fed} introduced task arithmetic in MaT-FL, enabling clients to train task-specific adapters and leverage server-side adapter fusion for optimizing multiple tasks. To further improve task disentanglement during model aggregation, they applied Neural Tangent Kernel (NTK) linearization over the model prior to training. Nevertheless, group-based MaT-FL approaches require the server to manage multiple models for each task group, assume task relationships are known in advance, and introduce significant client-side overhead, making scalability a challenge as the number of tasks increases. Training a single model across tasks offers considerable benefits: it reduces resource consumption and can enhance performance by leveraging diverse auxiliary tasks. While NTKFedAvg attempts to address these challenges, it focuses on a single task per client, whereas in practice, clients may hold multiple tasks.~\method~bridges this gap by training a unified model across all clients and tasks, dynamically building task correlations to enable knowledge transfer, and using task vector aggregation based on task similarity, removing the need for multiple server-side models. 

\section{Methodology}

\begin{figure*}[!t]
    \centering \small
    \includegraphics[width=0.75\textwidth]{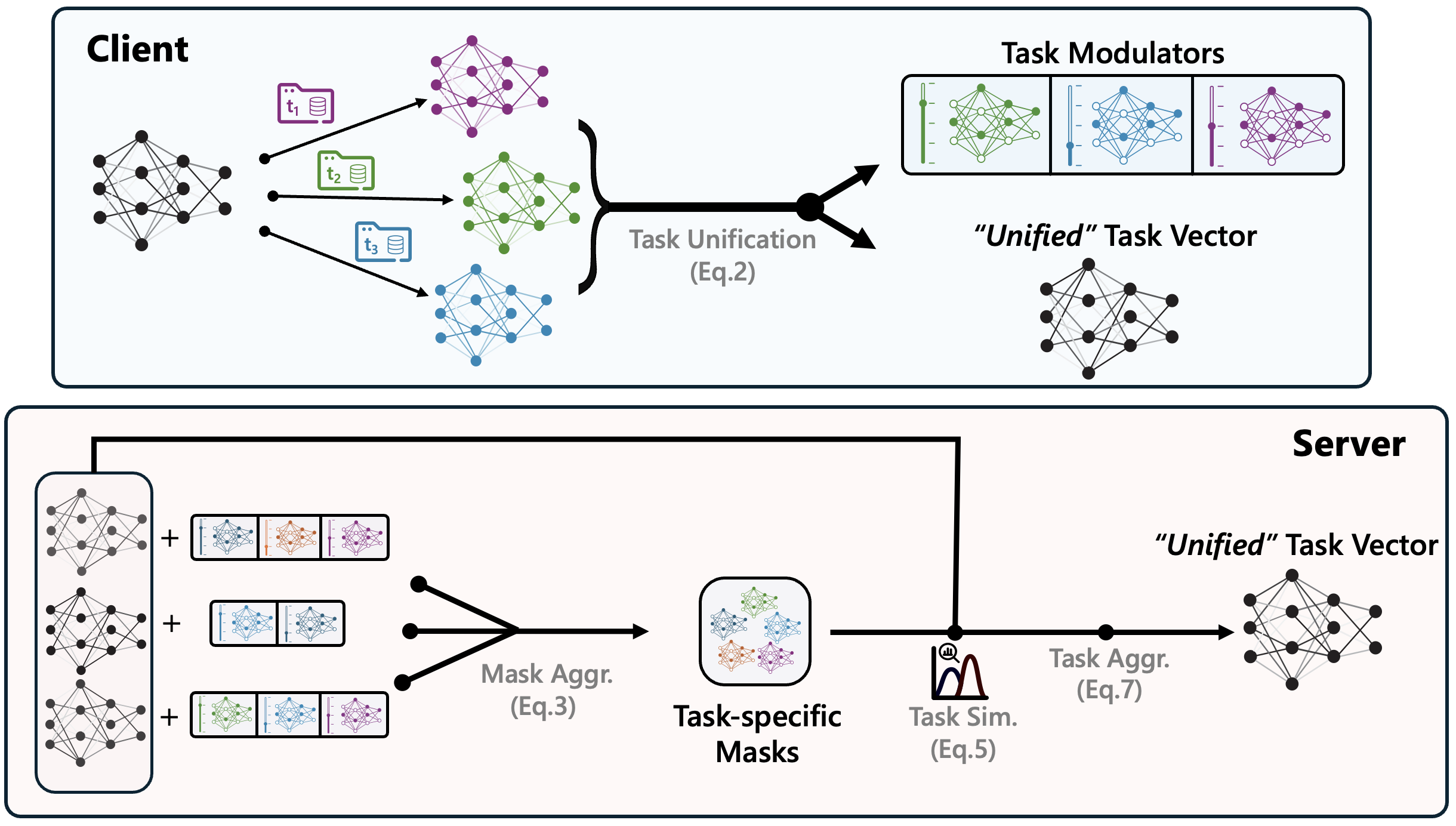}
    \vspace{-5pt}
    \caption{\small{\method's training process: Clients update their task vectors, and transmit their ``\textit{unified}'' task vector and task modulators to server, which constructs task masks, estimates task similarity via sign conflicts, and construct clients' task vectors using the top-$\kappa$ similar tasks.\label{fig:overview}}}
    \vspace{-10pt}
\end{figure*}

\subsection{Preliminaries}

\noindent \textbf{Task Arithmetic.} We denote with $\theta_{p} \in \mathbb{R}^d$ the pre-trained model weights of a FM. A task vector $\tau_{t}$ for task $t$ is defined as $\tau_{t} = \theta_{t}^{*} - \theta_{p}$, where $\theta_{t}^{*}$ refers to the fine-tuned model weights of task $t$. Essentially, $\tau_{t}$ can be seen as a representation of the direction (\textit{sign}) and amount (\textit{magnitude}) of movement relative to $\theta_{p}$ needed in the $d$-dimensional weights space to lead to a low loss region for the task $t$. Additionally, a scalar multiplication with $\lambda$ over $\tau_{t}$, i.e., $\tau_{new}^t = \lambda \cdot \tau_{t}$ yields models parameters along the training trajectory, with positive $\lambda$ values denoting task learning and negative $\lambda$ values indicating task negation (unlearning)~\cite{ortizjimenez2023task}. Hence, adding $K$ task vectors, a many-task (``\textit{merged}'') model can be obtained aiming to solve all $K$ tasks as follows:

\vspace{-10pt}
\begin{equation}\label{eqn:tv}
    \theta_{\text{m}} = \theta_{p} + \sum_{k=1}^{K} \lambda_k \cdot \tau_{k}
\end{equation}

By only adjusting $\lambda_k$, the performance across $K$ tasks can be improved. Building on the observations from~\cite{ortizjimenez2023task}, TIES~\cite{yadav2023tiesmerging} demonstrated that resolving sign conflicts before model merging enhances weight disentanglement (i.e., separating the influence of different task vectors), leading to improved task performance. Furthermore, re-scaling task vectors often yields better results than using unmodified vectors~\cite{yang2024adamergingadaptivemodelmerging,yu2024languagemodelssupermario}. \\ 

\noindent \textbf{Unified Task Vector.} Given a number of task vectors derived from a set of tasks, $t \in \mathcal{T}$, the ``\textit{task unification}''~\cite{huang2024emrmergingtuningfreehighperformancemodel} process begins by computing the aggregated sign vector across all task vectors, $\sigma$ = $\text{sgn}\left( \sum_{i=1}^{|\mathcal{T}|} \tau^i \right)$. Next, the ``\textit{unified}'' task vector is as follows:

\begin{equation} \label{eqn:tv_uni}
    \tau = \sigma \odot \mu ~,
\end{equation}

\noindent where \( \mu \) refers to the magnitude vector constructed by extracting the maximum absolute value from the task vectors whose signs align with \( \sigma \). This electing  procedure has shown to reserve the maximum amplitude and sign information shared by the task vectors, thereby maximally reducing interference~\cite{huang2024emrmergingtuningfreehighperformancemodel}.\\

\noindent \textbf{Weight disentanglement.} Weight disentanglement — the model’s ability to isolate the influence of different tasks on its weights — serves as a necessary condition for task arithmetic operations, and is an emerging property of pre-training~\cite{ortizjimenez2023task}. Since the sign of each weight indicates the direction that minimizes loss for a given task~\cite{ortizjimenez2023task,yadav2023tiesmerging,muhamed2024fed}, resolving sign conflicts among task vectors can promote weight disentanglement, leading to improved individual task performance~\cite{yadav2023tiesmerging}. Therefore, in the context of the same pre-trained model, a high count of sign conflicts between two task vectors implies opposing influences on model weights, suggesting potential interference in their joint training. 

\subsection{\method~:Efficient MaT-FL with Task Arithmetic.}

\noindent \textbf{Overview.} Here, we present the general \method~training pipeline (see Fig.~\ref{fig:overview}). Clients initialize the same model \( f \), parameterized by the pre-trained weights \( \theta_p \), and train on their local datasets \( D^t_n \) to generate individual task vectors \( \tau^t_n \), which are combined into a unified task vector \( \tau_n \). To account for the changes from \( \tau^t_n \) to \( \tau_n \), clients create lightweight modulators — binary masks \( m^i_n \) and scaling values \( \lambda^i_n \) — which are transmitted along with \( \tau_n \). Upon receiving client updates, for each task, the server computes an average task mask, \( \hat{m}^t \), to capture key areas in the task vectors, then performs task-specific aggregation to combine updates from clients sharing the task, followed by cross-task aggregation to enable knowledge transfer between related tasks. By averaging these two, the server creates the updated task vectors to compute unified task vectors and modulators for each client based on their assigned tasks, which then transmits to client. Next, clients use these modulators to adjust the unified task vector and derive the updated task vectors, marking the next federated round. \\

\noindent \textbf{Notation.} We use the following notation in the rest of the paper. Let \( \mathcal{N} \) refer to a set of clients, each assigned a set of $k$ tasks \( \mathcal{T}_n \) ($k_n=|\mathcal{T}_n|$), which may overlap across clients (i.e., the unique set of tasks is given by \( \mathcal{T} = \bigcup_{n=1}^{|\mathcal{N}|} \mathcal{T}_n \)). Each task $t \in \mathcal{T}_n$ is associated with a locally stored dataset $D^t_n$, containing $|D^t_n|$ samples. $A \in \{0, 1\}^{N \times T}$ denotes a binary matrix representing \textit{task-to-client allocations}, where $A(n,t)$=$1$ if client $n$ holds task $t$, and $0$ otherwise. The neural network is denoted by $f$ and parameterized by weights $\theta$, where $\theta_p$ represents its pre-trained weights, and $\theta^*_t$ denotes its optimized weights after local training on task $t$. For each task $t$, its task vector is given by $\tau^t = \theta^*_t - \theta_p$, where $\sigma^t = \text{sgn}(\tau^t)$ denotes the sign vector of $\tau_t$, indicating the direction of movement (positive or negative) relative to $\theta_p$.

\noindent \textbf{Local Training with many-tasks.} In a given round \( r \in R \), the $n$-th client trains individually across the set of $k_n$ locally stored tasks, and derives a set of local task vectors, one for each task \( t \in \mathcal{T}_n \). Next, the ``\textit{task unification}'' process is performed to derive the client's ``\textit{unified}'' task vector, $\tau_n = \sigma_n \odot \mu_n $.

As a single task vector cannot capture all task-specific weights, leading to performance degradation, we introduce \textit{lightweight} task-specific modulators that refine \( \tau_n \) to approximate task-specific vectors, as in~\cite{huang2024emrmergingtuningfreehighperformancemodel}. Specifically, we construct a set of task-specific masks, \( \mathcal{M}_n = \{ m^i_n \}^{k_n}_{i=1} \), where each mask is defined as \( m^i_n = (\tau^i_n \odot \tau_n > 0) \) for each task \( t \in \mathcal{T}_n \); thus, aligning the direction of \( \tau_n \) to each task vector. Similarly, to account for the magnitude shift between \( \tau_n \) and individual task vectors, we introduce task-specific scaling parameters \( \lambda_n = \left\{ \lambda^i_n \right\}_{i=1}^{k_n} \), where \( \lambda^i = \frac{\sum |\tau^i_n|}{\sum |m^i_n \odot \tau_n|} \). \\

\noindent \textbf{Many-tasks Aggregation.} Once the local training at round \( r \) is completed, the server holds each client's ``\textit{unified}'' task vector and task-specific modulators (i.e., masks and re-scaling parameters). We aim to aggregate task vectors among clients holding a given task, while enabling \textit{dynamic} knowledge transfer across similar tasks. Thus, we propose a simple-yet-effective \textit{many-tasks aggregation} scheme tailored for MaT-FL, incorporating both task-specific and cross-task aggregation mechanisms. \\

\noindent \textit{Task-specific Aggregation.} For each task $t$, we compute the average task-specific mask $\hat{m}^t$ across the clients that hold this task, defined as $\mathcal{N}^t = \{ n \in \mathcal{N} \mid A(n,t) = 1 \}$, by performing the following element-wise operation on its elements:

\vspace{-10pt}
\begin{equation}\label{eqn:avg_task_mask}
    \resizebox{0.43\textwidth}{!}{
    $
    [\hat{m}^t]_j = 
    \begin{cases} 
        1 & \text{if } \alpha^t_j \geq \rho \\ 
        \alpha^t_j & \text{otherwise}
    \end{cases},~\text{where}~\alpha^t_j = \left| \frac{1}{|\mathcal{N}^t|} \sum\limits_{n \in \mathcal{N}^t} \text{sgn}(m^t_n \odot \tau_n) \right|
    $
    }
\end{equation}

\noindent where \( [\hat{m}^t]_j \) represents the \( j \)-th element of \( \hat{m}^t \), and \( \rho \)\footnote{\( \rho = 0.4 \) following~\cite{tenison2023gradientmaskedaveragingfederated}} is a threshold that determines the significance of each parameter for task \( t \). 

As demonstrated in~\cite{tenison2023gradientmaskedaveragingfederated}, the agreement score \( \alpha \) is closely related to client heterogeneity, with highly heterogeneity among clients resulting in \( \alpha \) close to $0$. As clients optimize their task vectors across distinct sets of tasks, the impact on each task-specific vector entry varies due to unique interference from their remaining tasks, exacerbating their heterogeneity. By adjusting the magnitude of updates based on the agreement score $\alpha$, we effectively reduce task interference, enhancing task separation; thus, promoting weight disentanglement during FL training stage. Using the $\hat{m}^t$, we then compute the average task-specific vectors across clients as:

\vspace{-10pt}
\begin{equation}\label{eqn:same_task_agg}
    \hat{\tau}^t = \sum_{n \in \mathcal{N}^t} \gamma_{n}^t \cdot \lambda_{n}^t \cdot \hat{m}^t \odot \tau_{n}^{t}~,
\end{equation}

\noindent where \( \lambda_{n'}^t \) is the task-specific re-scaler for client \( n' \) and task \( t \), \( \hat{m}_t \) is the aggregated task-specific mask for task \( t \) (Eq.~\ref{eqn:avg_task_mask}), and \( \gamma_n^t \) represents clients' weight in the aggregation based on their respective dataset size (i.e., \( \gamma_n^t = |D^t_n| / \smash{\sum_{n \in \mathcal{N}^t} |D^t_n|} \)), similar to FedAvg~\cite{mcmahan2017communication}. In essence, \( \hat{\tau}^t \) captures the task-specific information aggregated across clients in the FL process. \\

\noindent \textit{Cross-task Aggregation.} Apart from capturing knowledge among clients holding task $t$, leveraging information from similar tasks can enhance task-specific performance, especially for data-scarce tasks~\cite{muhamed2024fed}. To identify redundancies that facilitate joint learning and enhance individual task performance, we first examine task similarity—the relationships among tasks~\cite{zamir2018taskonomydisentanglingtasktransfer}. Most task similarity metrics evaluate knowledge transfer benefits based on task-specific data~\cite{gholami2023etranenergybasedtransferabilityestimation,bao2022informationtheoreticapproachtransferabilitytask} or labels~\cite{ding2022pactranpacbayesianmetricsestimating}; yet, access to these resources is often limited in FL. Instead, as a high count of sign conflicts between task vectors for the same pre-trained model indicates potential interference in joint training~\cite{yadav2023tiesmerging}, we construct a task similarity matrix from sign conflicts among aggregated task-specific vectors. Specifically, the similarity between tasks  $t$  and  $t'$  is defined as follows:

\vspace{-10pt}
\begin{equation}\label{eqn:sim_matrix}
    \resizebox{0.43\textwidth}{!}{
    $
        S(t,t') = \frac{1}{2} \left( \frac{1}{d} \sum\limits_{i=1}^{d} \left( \text{sgn}\left([\hat{\tau}^t]_i\right) \cdot \text{sgn}\left([\hat{\tau}^{t'}]_i\right)\right) + 1 \right)
    $
    }
\end{equation}

\noindent where \( d \) is the dimension of the task vectors. Note that \( S \in [0, 1]\), with a higher score indicating better alignment between tasks. Next, for a given task \( t \), we derive the set of top-$\kappa$ similar tasks, $\mathcal{Z}^t = [ \{ t' \in \mathcal{T} \mid S(t, t') > \epsilon \} ]_{\le \kappa}$\footnote{\( \epsilon = 0.5 \) filters out low-similarity tasks, and $[\cdot]_{\le \kappa}$ refer to top-$\kappa$.}, and use it to compute the average cross-task vectors as follows:

\begin{equation}\label{eqn:cross_task_agg}
    \tilde{\tau}^t = \sum_{t' \in \mathcal{Z}^t} S \left(t,t'\right) \cdot \hat{m}^t \odot \hat{\tau}^{t'}~,
\end{equation}

\noindent Note that we use $\hat{m}^{t}$, rather than $\hat{m}^{t'}$, to adapt the $t$-th task's vector based on the aggregated task vectors of task $t'$, effectively enabling knowledge transfer across tasks. \\

\noindent \textit{Forming clients updates.}  Given Eq.~\ref{eqn:same_task_agg}~and~\ref{eqn:cross_task_agg}, we can compute the aggregated task-specific vector for task $t$ on round $r$, as:

\vspace{-10pt}
\begin{equation}\label{eqn:task_agg}
    \resizebox{0.445\textwidth}{!}{
    $
        \tau^{t,r+1} = \hat{\tau}^{t,r} + \tilde{\tau}^{t,r} = \underbrace{\sum_{n \in \mathcal{N}^t,r} \gamma_{n}^t \cdot \lambda_{n}^{t,r} \cdot \hat{m}^{t,r} \odot \tau_{n}^{t,r}}_{\text{same-task}}~~+~~\underbrace{\sum_{t' \in \mathcal{Z}^{t,r}} S \left(t,t'\right) \cdot \hat{m}^{t,r} \odot \hat{\tau}^{t',r}}_{\text{cross-tasks}}
    $
    }
\end{equation}

\begin{figure}[!t]
    \centering
    \begin{subfigure}[b]{0.155\textwidth}
        \centering
        \includegraphics[width=\textwidth]{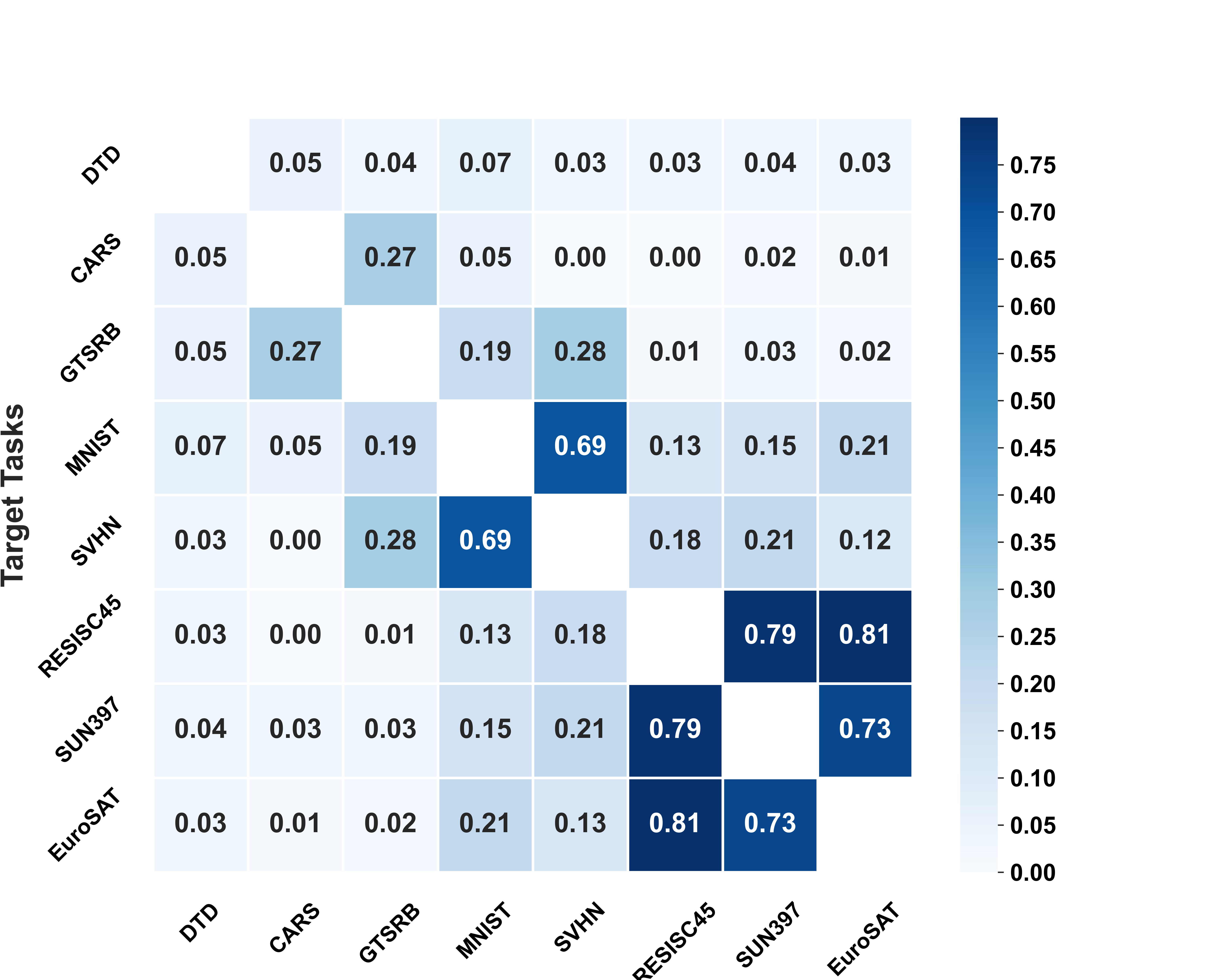}
        \caption{Cosine Similarity} \label{fig:cosine}
    \end{subfigure}
    \begin{subfigure}[b]{0.155\textwidth}
        \centering
        \includegraphics[width=\textwidth]{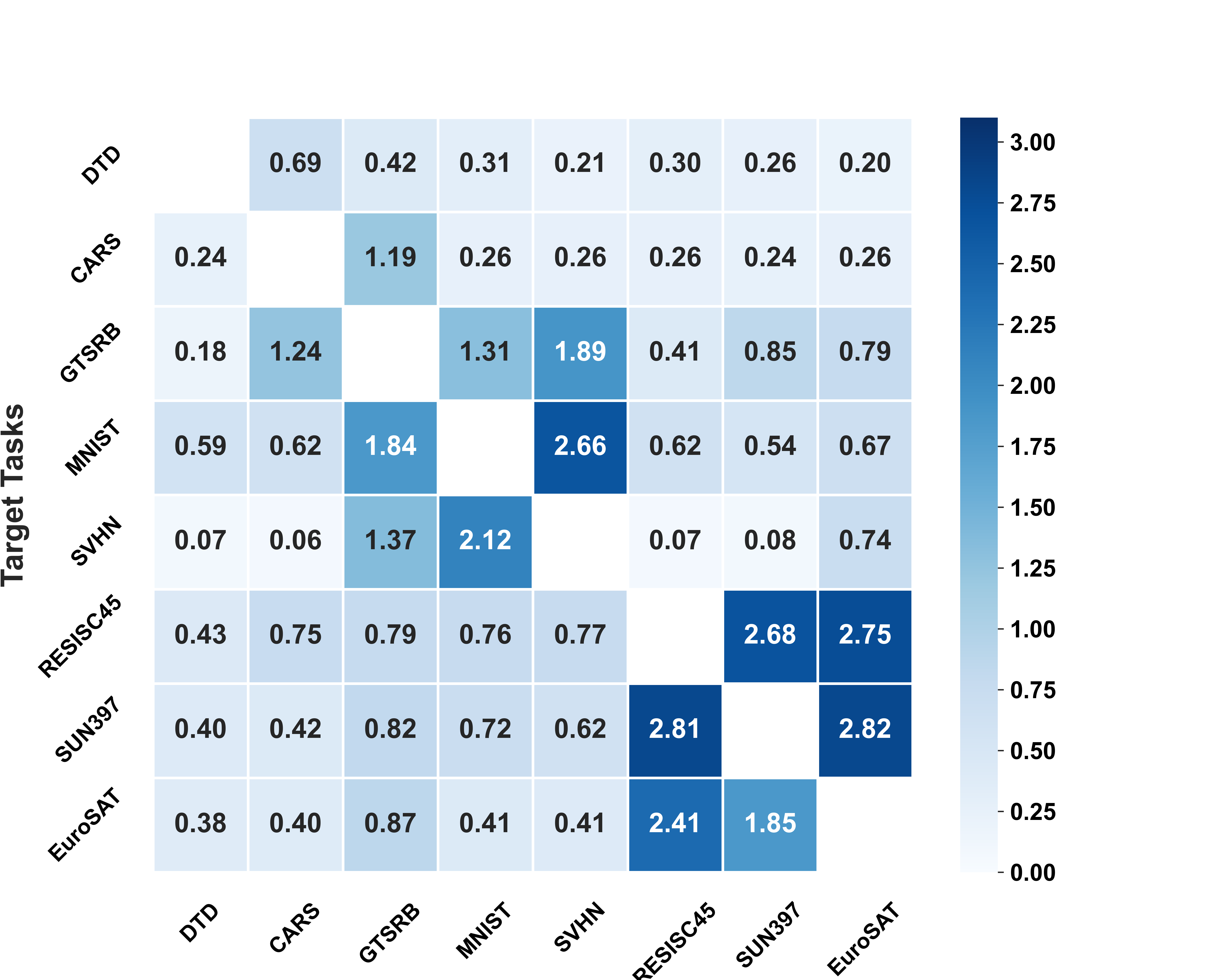}
        \caption{LogME} \label{fig:logme}
    \end{subfigure}
    \begin{subfigure}[b]{0.155\textwidth}
        \centering
        \includegraphics[width=\textwidth]{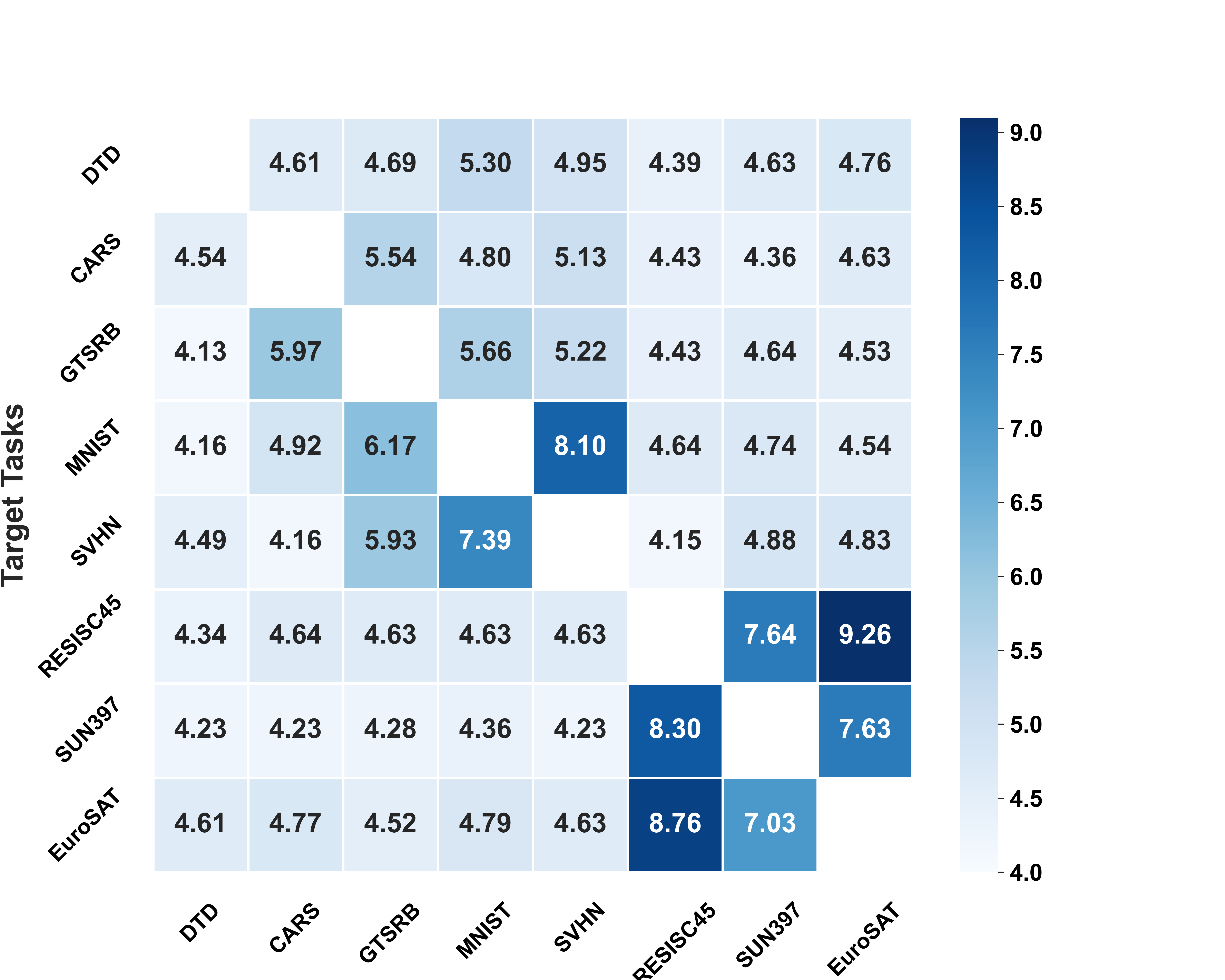}
        \caption{E-Trans} \label{fig:etrans}
    \end{subfigure}
    \begin{subfigure}[b]{0.155\textwidth}
        \centering
        \includegraphics[width=\textwidth]{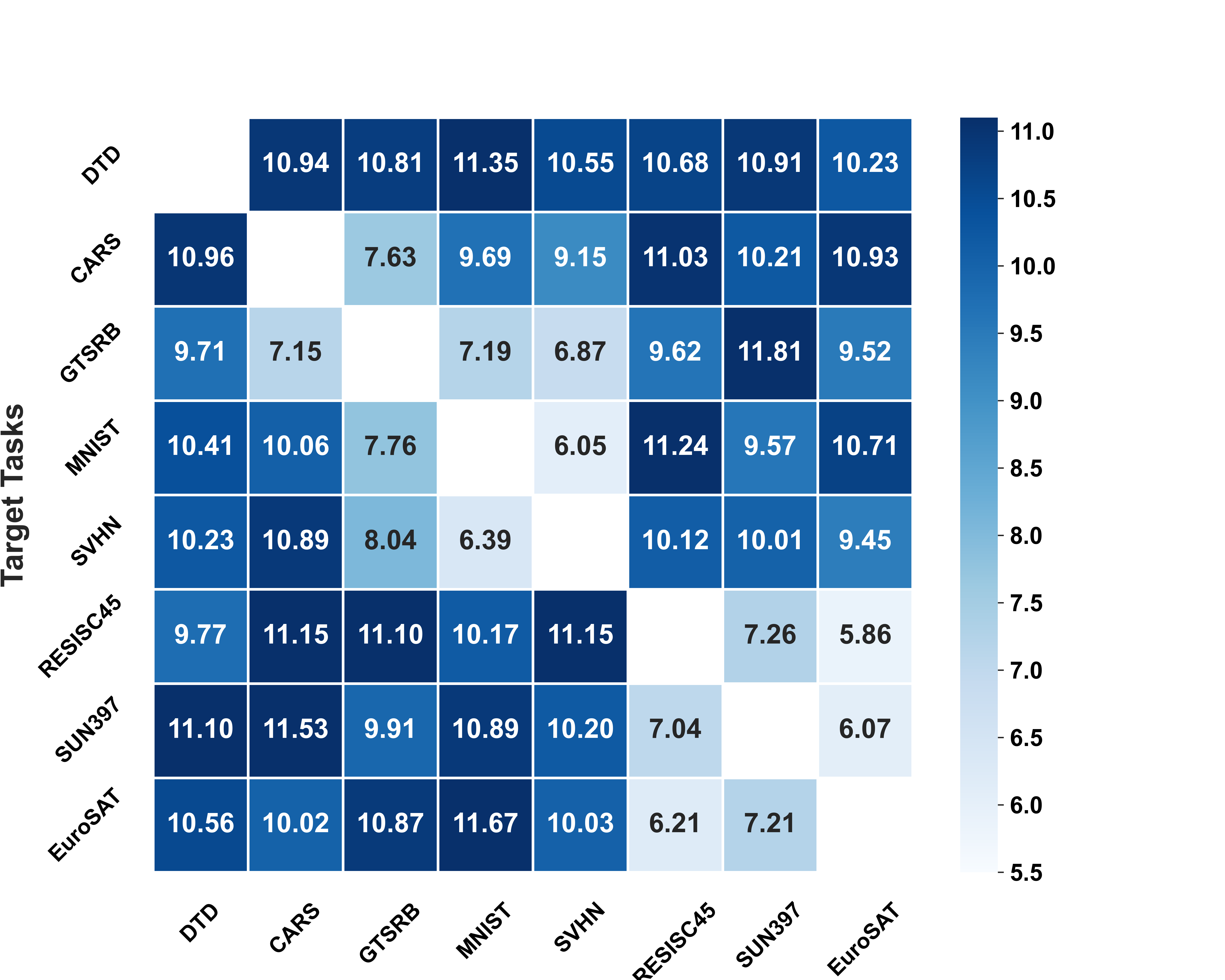}
        \caption{WTE} \label{fig:wte}
    \end{subfigure}
    \begin{subfigure}[b]{0.155\textwidth}
        \centering
        \includegraphics[width=\textwidth]{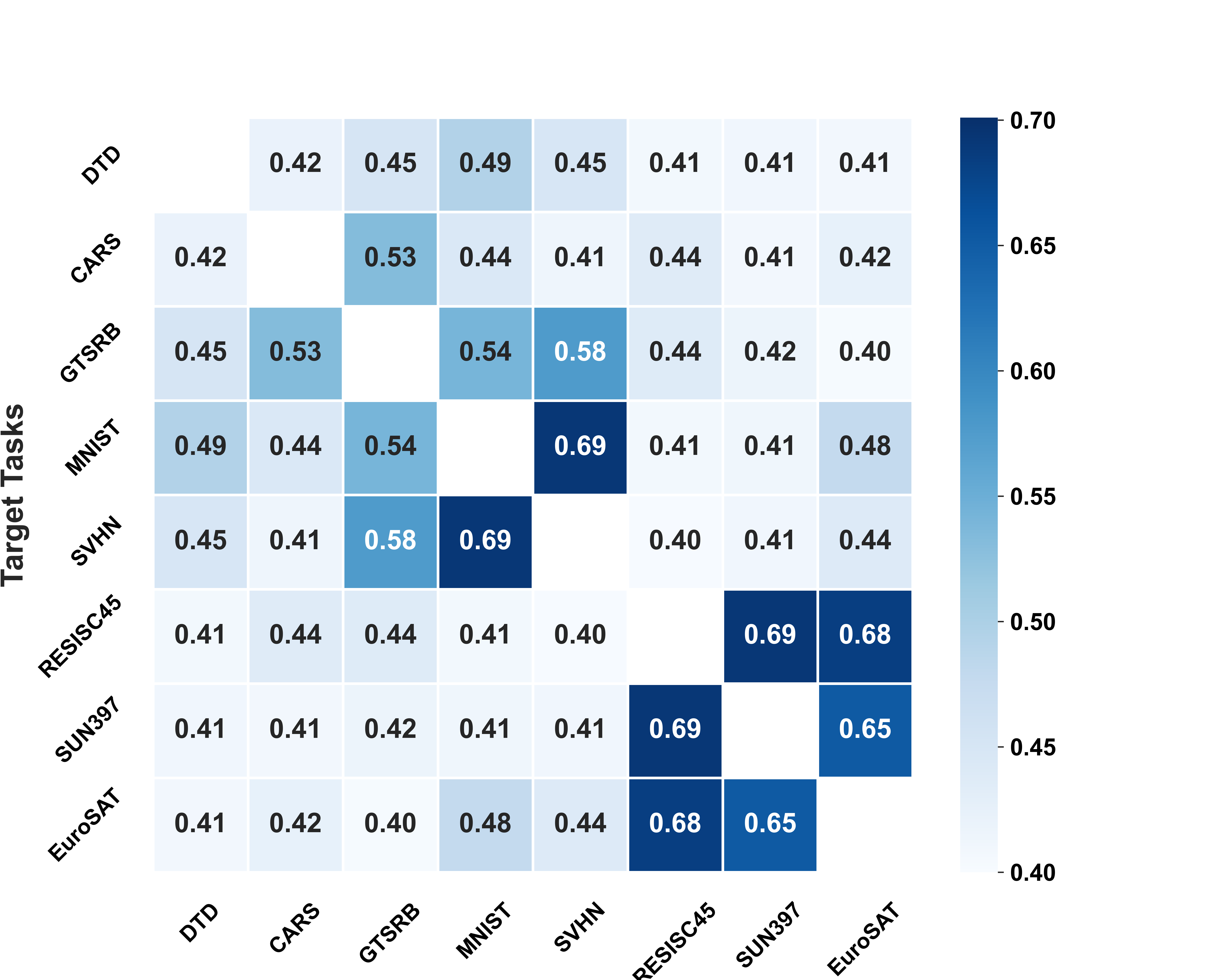}
        \caption{FIM} \label{fig:fim}
    \end{subfigure}
    \begin{subfigure}[b]{0.155\textwidth}
        \centering
        \includegraphics[width=\textwidth]{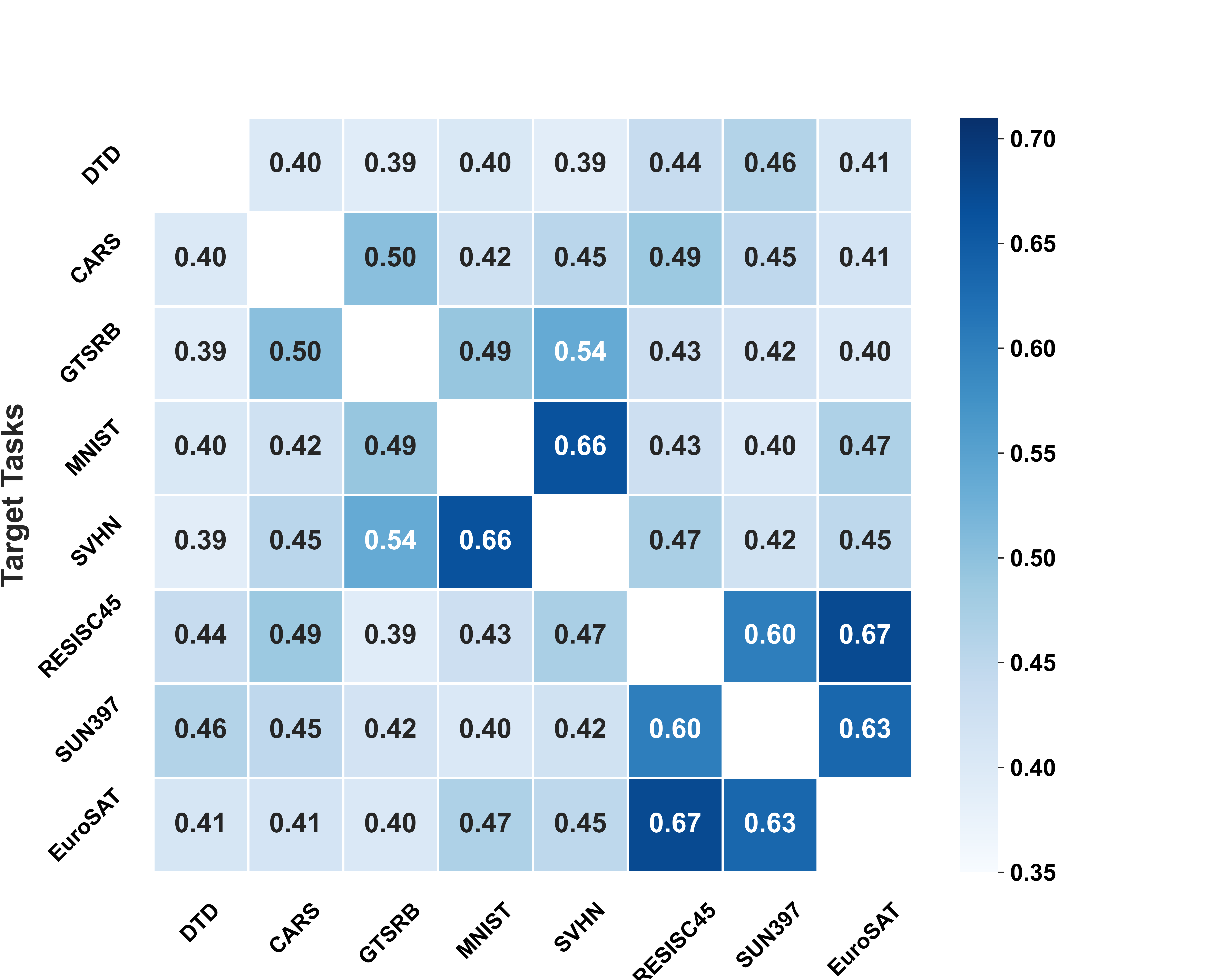}
        \caption{Sign (\textbf{Ours})} \label{fig:sing}
    \end{subfigure}
    \caption{Comparison of sign vectors vs. state-of-the-art transferability estimation metrics across $8$ datasets. In all metrics except WTE, higher values correspond to higher correlation.}
    \label{fig:sim_matrices}
    \vspace{-10pt}
\end{figure}

For the set of tasks assigned to each clients, $\mathcal{T}_n$, we can now utilize the newly computed aggregated task-specific vector to derive their updated ``\textit{unified}'' task vector, $\tau^{r+1}_n$, and subsequently compute the set of task-specific masks, \( \mathcal{M}^{r+1}_n = \{ (\tau^{i,r+1} \odot \tau^{r+1}_n > 0) \}^{k_n}_{i=1} \), and scaling parameters \( \lambda^{r+1}_n = \left\{ \frac{\sum |\tau^{i,r+1}_n|}{\sum |m^{i,r+1}_n \odot \tau^{r+1}_n|} \right\}_{i=1}^{k_n} \). The client’s ``\textit{unified}'' task vector, $\tau^{r+1}_n$, and task-specific modules, $\mathcal{M}^{r+1}_n$  and $\lambda^{r+1}_n$, are then transmitted to the client-side for subsequent training. To train task $t$, the client must modulate the ``\textit{unified}'' vector as $\dot{\tau}^{r+1}_t = \lambda^{t,r+1}_n \cdot m^{t,r+1}_n \odot \tau^{r+1}_n$, and injects it to the pre-trained model (i.e., $\theta^{r+1}_t = \theta_p + \dot{\tau}^{r+1}_t$). In contrast to personalized FL or group-based MaT-FL methods that require storing task-specific heads or group models, our design remains stateless — the server only handles task-level aggregation and does not retain client-specific weights across rounds. 

\section{Performance evaluation}

\noindent \textbf{Datasets.} We evaluate~\method's performance across $30$ vision classification datasets. Specifically, we consider $2$ model merging benchmarks: (i) an \textit{$8$-task benchmark} comprising SUN397~\cite{sun397}, Cars~\cite{cars}, RESISC45~\cite{resisc45}, EuroSAT~\cite{eurosat}, SVHN~\cite{svhn}, GTSRB~\cite{gtsrb}, MNIST~\cite{mnist}, and DTD~\cite{dtd}; and (ii) a large-scale \textit{$30$-task benchmark} encompassing diverse vision tasks as proposed in~\cite{huang2024emrmergingtuningfreehighperformancemodel}. \\

\noindent \textbf{Baselines.} We compare~\method~against centralized LoRA PEFT~\cite{hu2021loralowrankadaptationlarge}, traditional FL methods (FedAvg~\cite{mcmahan2017communication}, FedProx~\cite{li2020federatedoptimizationheterogeneousnetworks}), group-based approaches (\textit{MaT-FL}~\cite{10208649}), personalized FL approaches (FedPer~\cite{arivazhagan2019federatedlearningpersonalizationlayers}) and related task-arithmetic methods like NTKFedAvg~\cite{muhamed2024fed}. In FedPer, the last ViT block and classifier are used as personalized layers. We compare all baselines in terms of model performance (i.e., accuracy on test set) and client's bitrate requirements (e.g. communicated bits per task in a round~\textendash~$\mathrm{bpr}$). All experiments utilize ViT-B/32~\cite{dosovitskiy2021imageworth16x16words} pre-trained on ImageNet-21k~\cite{5206848} and employ LoRA~\cite{hu2021loralowrankadaptationlarge} with a rank of $16$ for PEFT, reporting test set accuracy. \\

\noindent \textbf{FL Settings.} We perform our FL simulations using Flower~\cite{beutel2020flower} with key parameters being number of clients ($N$), rounds ($R$), local epochs ($E$), clients' participation rate ($\xi$), number of tasks ($T$), class concentration across clients ($\zeta_{c}$), and task concentration across clients ($\zeta_{t}$). We used a Dirichlet distribution, denoted as Dir($\alpha$), for both data and task splitting, where $\alpha$ represents the concentration parameter, following~\cite{li2021federatedlearningnoniiddata}. We simulate non-IID settings with $\alpha$=$0.1$ ($\zeta_c$=$\zeta_t$=$0.1$) and train federated models for $100$ rounds ($R$=$100$) with $30$ clients ($N$=$30$), each performing one local training step per round ($E$=$1$). Note that for $\xi$\textless$1$, clients are randomly selected. \\

\begin{figure}[!t]
    \small \centering
    \includegraphics[width=0.435\textwidth, trim=10 10 10 10, clip]{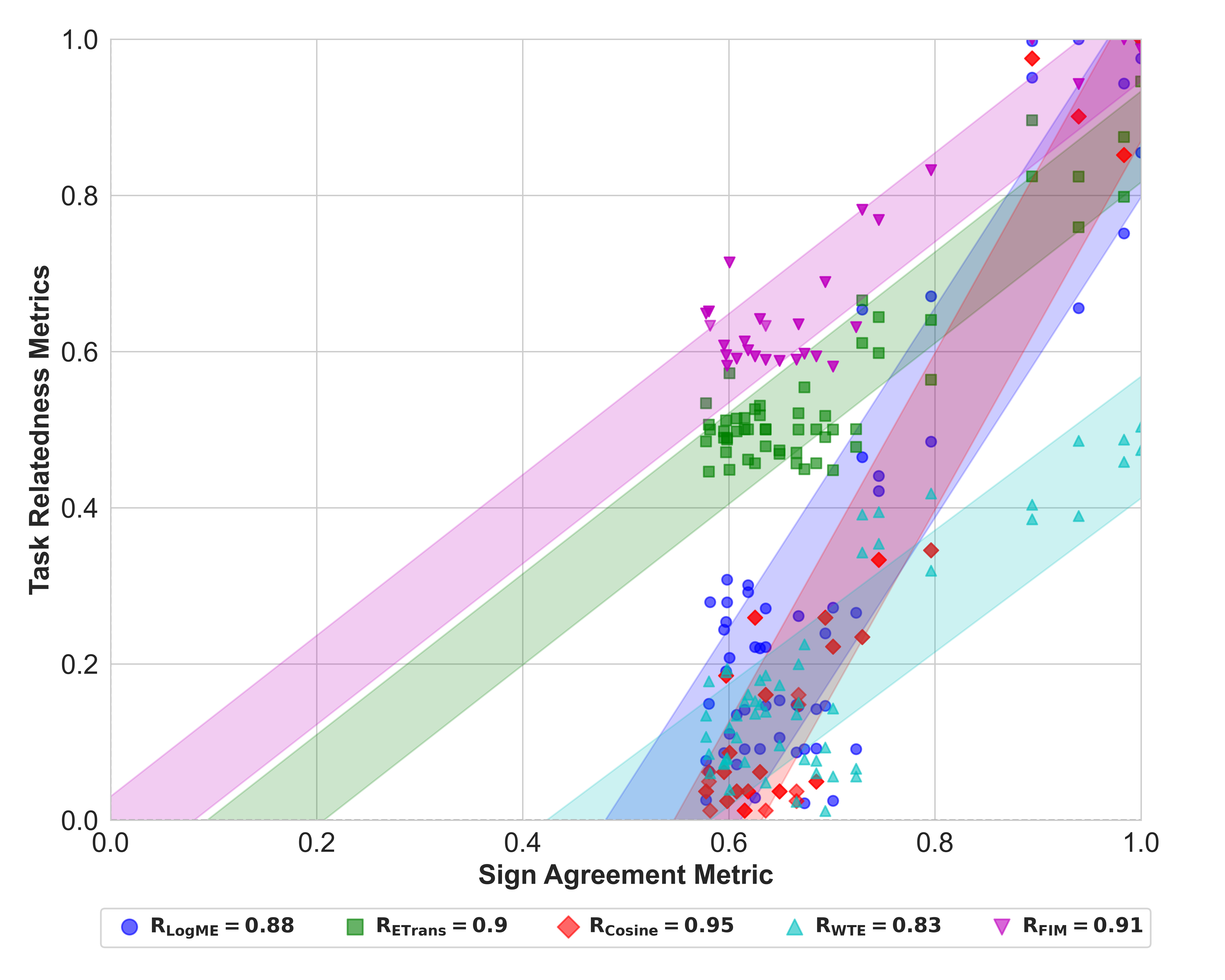}
    \vspace{-5pt}
    \caption{\small{Pearson Correlation between sign-based and state-of-the-art task relatedness metrics across $8$ datasets.}} \label{fig:corr}
    \vspace{-10pt}
\end{figure}

\noindent \textbf{Task Relatedness.} We investigate the effectiveness of sign agreement among task vectors as a metric for measuring task similarity across datasets. To evaluate this, we conduct experiments with $8$ vision classification tasks, comparing our sign-conflict approach against well-established task similarity metrics, including cosine similarity of weights~\cite{vu2022spotbetterfrozenmodel}, E-Trans~\cite{gholami2023etranenergybasedtransferabilityestimation}, FIM~\cite{Amari1998NaturalGW}, and WTE distance~\cite{LIU2025106796}. As shown in Fig.\ref{fig:sim_matrices}, our sign-conflict method effectively identifies similar task groups among the datasets. Furthermore, Fig.\ref{fig:corr} highlights a strong Pearson correlation (\textgreater 0.8) between the sign-conflict-based and all other task similarity metrics.

\subsection{Results}\label{ssec:results}


\begin{table*}[!t]
    \centering \small
    \caption{\small{Performance evaluation of ViT-B/32 models on $8$ tasks in a \textbf{\textit{single-task per client}} setting. Reported bitrates ($\mathrm{bpt}$) are in millions. Federated parameters are set to $N$=$30$, $E$=$1$, $\zeta_t$=$0.0$, $\xi$=$0.2$ and $R$=$100$ (except~\textit{MaT-FL}~where~$R$=$300$).}}
    \label{tab:stask_perf}
    \vspace{-5pt}
    \resizebox{0.85\textwidth}{!}{%
        \begin{tabular}{lcccccccccc}
        \toprule
        \textbf{Methods} & \textbf{SUN397} & \textbf{Cars} & \textbf{RESISC45} & \textbf{EuroSAT} & \textbf{SVHN} & \textbf{GTSRB} & \textbf{MNIST} & \textbf{DTD} & \textbf{Avg. Acc} & \textbf{Bitrate ($\mathrm{bpt}$ - M)}  \textcolor{green}{$\downarrow$} \\ 
        \midrule
        Individual (Centralized)                                            & 74.81 & 76.93 & 96.05 & 99.43 & 97.05 & 98.57 & 99.67 & 79.21 & 90.21 & \textendash \\ 
        \midrule
        FedAvg~\cite{mcmahan2017communication}                              & 62.84 & 60.77 & 71.08 & 72.58 & 68.72 & 65.06 & 65.47 & 50.28 & 64.60 & 6.32 \\ 
        NTK-FedAvg \cite{muhamed2024fed}                                    & 63.08 & 59.22 & 71.98 & 72.37 & 67.34 & 66.13 & 67.04 & 54.97 & 65.13 & 6.32 \\ 
        FedProx~\cite{li2020federatedoptimizationheterogeneousnetworks}     & 66.58 & 63.64 & 74.95 & 76.26 & 70.02 & 62.86 & 67.65 & 55.16 & 67.14 & 6.32 \\ 
        \midrule
        FedPer~\cite{arivazhagan2019federatedlearningpersonalizationlayers} & 68.57 & \textbf{69.39} & 88.37 & 90.33 & 90.28 & 91.53 & 92.44 & 67.98 & 82.11 & \textbf{5.36} \\ 
        \textit{MaT-FL}~\cite{10208649}                                     & 67.02 & 65.19 & 85.13 & 82.68 & 85.90 & 80.16 & 87.08 & 60.12 & 76.66 & 6.32 \\ 
        \midrule
        \textbf{\method}~(\textbf{Ours})                                      & \textbf{71.61} & 69.17 & \textbf{91.27} & \textbf{92.59} & \textbf{93.59} & \textbf{94.10} & \textbf{96.16} & \textbf{70.09} & \textbf{84.32} & 6.32 \\ 
        \bottomrule
        \end{tabular}%
    }
    \vspace{-10pt}
\end{table*}

\begin{table*}[!t]
    \centering \small
    \caption{\small{Performance evaluation of ViT-B/32 models on $8$ tasks in a \textbf{\textit{multiple-task per client}} setting. Reported bitrates ($\mathrm{bpt}$) are in millions. Federated parameters are set to $N$=$30$, $E$=$1$, $\zeta_t$=$0.5$, $\xi$=$0.2$ and $R$=$100$ (except~\textit{MaT-FL}~where~$R$=$300$).}}\label{tab:multitask_perf}
    \vspace{-5pt}
    \resizebox{0.85\textwidth}{!}{%
        \begin{tabular}{lcccccccccc}
        \toprule
        \textbf{Methods} & \textbf{SUN397} & \textbf{Cars} & \textbf{RESISC45} & \textbf{EuroSAT} & \textbf{SVHN} & \textbf{GTSRB} & \textbf{MNIST} & \textbf{DTD} & \textbf{Avg. Acc} & \textbf{Bitrate ($\mathrm{bpt}$ - M)} \textcolor{green}{$\downarrow$} \\ 
        \midrule
        Individual (Centralized)                                            & 74.81 & 76.93 & 96.05 & 99.43 & 97.05 & 98.57 & 99.67 & 79.21 & 90.21 & \textendash \\ 
        \midrule
        FedAvg~\cite{mcmahan2017communication}                              & 50.21 & 48.50 & 56.97 & 57.86 & 55.19 & 52.04 & 52.26 & 40.01 & 51.63 & 20.43 \\ 
        NTK-FedAvg \cite{muhamed2024fed}                                    & 55.37 & 53.27 & 62.30 & 63.62 & 58.56 & 52.79 & 56.27 & 46.28 & 56.06 & 20.43 \\ 
        FedProx~\cite{li2020federatedoptimizationheterogeneousnetworks}     & 57.64 & 53.84 & 63.26 & 64.44 & 59.17 & 52.90 & 56.76 & 42.54 & 56.32 & 20.43 \\ 
        \midrule
        FedPer~\cite{arivazhagan2019federatedlearningpersonalizationlayers} & 41.29 & 38.26 & 57.01 & 61.20 & 57.06 & 55.06 & 36.32 & 40.74 & 48.37 & 17.16 \\ 
        \textit{MaT-FL}~\cite{10208649}                                     & 61.52 & 56.67 & 76.48 & 73.95 & 76.96 & 69.67 & 74.64 & 50.88 & 67.60 & 20.43 \\ 
        \midrule
        \textbf{\method}~(\textbf{Ours})                                      & \textbf{66.40} & \textbf{63.46} &\textbf{ 86.36} & \textbf{87.26} & \textbf{88.15} & \textbf{88.51} & \textbf{91.08} & \textbf{64.69} & \textbf{79.47} & \textbf{8.04} \\ 
        \bottomrule
        \end{tabular}%
    }
    \vspace{-10pt}
\end{table*}

\noindent \textbf{Single-task Clients.} We explore a ``\textit{simplified}" version of MaT-FL, where each client holds a single task. Specifically, we perform experiments with $30$ clients ($N$=$30$) under low client participation rate ($\xi$=$0.2$), ensuring no overlap among client's tasks ($\zeta_t$=$0.0$) in the $8$-task benchmark. We report our findings in Table~\ref{tab:stask_perf}, from which we observe that~\method~achieves the highest average accuracy (84.32\%) among all baselines~\textendash~less than a 6\% gap to individual task fine-tuning performance. Traditional FL methods like FedAvg, NTK-FedAvg, and FedProx, with average accuracies of 64.60\%, 65.13\%, and 67.14\%, respectively, struggle to merge task-specific updates effectively, even in single-task settings. The Neural Tangent Kernel (NTK) linearization provides minimal improvements over standard FedAvg in task disentanglement (less than 1\%). While MaT-FL improves performance by mitigating task conflicts through cosine similarity-based grouping aggregation, it exhibits the slowest convergence among all baselines due to restricted client interactions. FedPer's task-specific personalization layers deliver strong performance (82.11\%) in single-task settings~\textendash~as they are explicitly tailored to individual tasks. In contrary,~\method~outperforms FedPer by 2.21\% on average, requiring only a marginal increase in bitrate. These results highlight~\method's ability to mitigate task interference via task-specific masks in aggregation, ensuring task disentanglement and superior performance without added communication overhead compared to standard FL regimes. \\



\begin{figure*}[!t]
    \small \centering
    \includegraphics[width=0.9\textwidth]{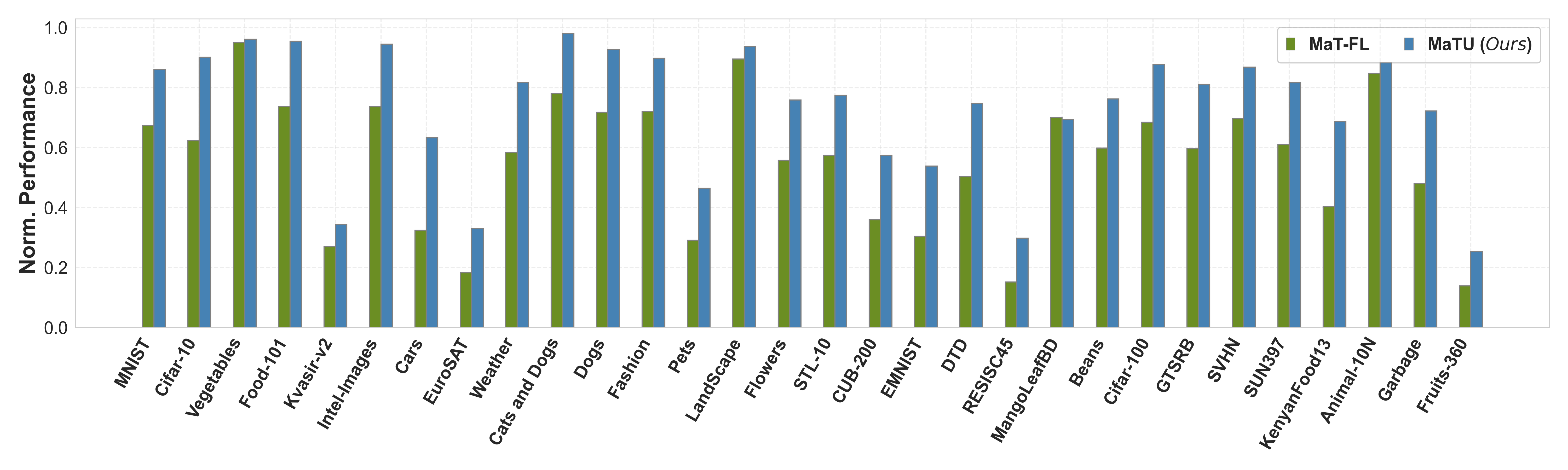}
    \vspace{-15pt}
    \caption{\small{Performance evaluation of ViT-B/32 models on the $30$-task benchmark. We report test set accuracy normalized to individual task fine-tuning performance. Federated parameters are set to $N$=$30$, $R$=$300$, $E$=$1$, $\zeta_t$=$0.2$, and $\xi$=$1.0$.}} \label{fig:30_vit_tasks}
    \vspace{-10pt}
\end{figure*}

\noindent \textbf{Multiple-task Clients.} We now explore a generalized MaT-FL scenario, where each client can hold an arbitrary (\textgreater$1$) set of tasks. For this, we perform experiments with $30$ clients ($N$=$30$) in both $8$-task and $30$-task benchmarks. Here, task conflicts are expected to intensify both within and across clients, as no restrictions are placed on task assignments (i.e., task groups are formed randomly given $\zeta_t$), allowing conflicting tasks to exist on a client. We report our results for both benchmarks in Table~\ref{tab:multitask_perf} and Fig.~\ref{fig:30_vit_tasks}. \\

\begin{figure}[!t]
    \centering \small
    \begin{minipage}{0.495\linewidth}
        \begin{subfigure}[b]{1.0\linewidth}
            \centering
            \includegraphics[width=\textwidth]{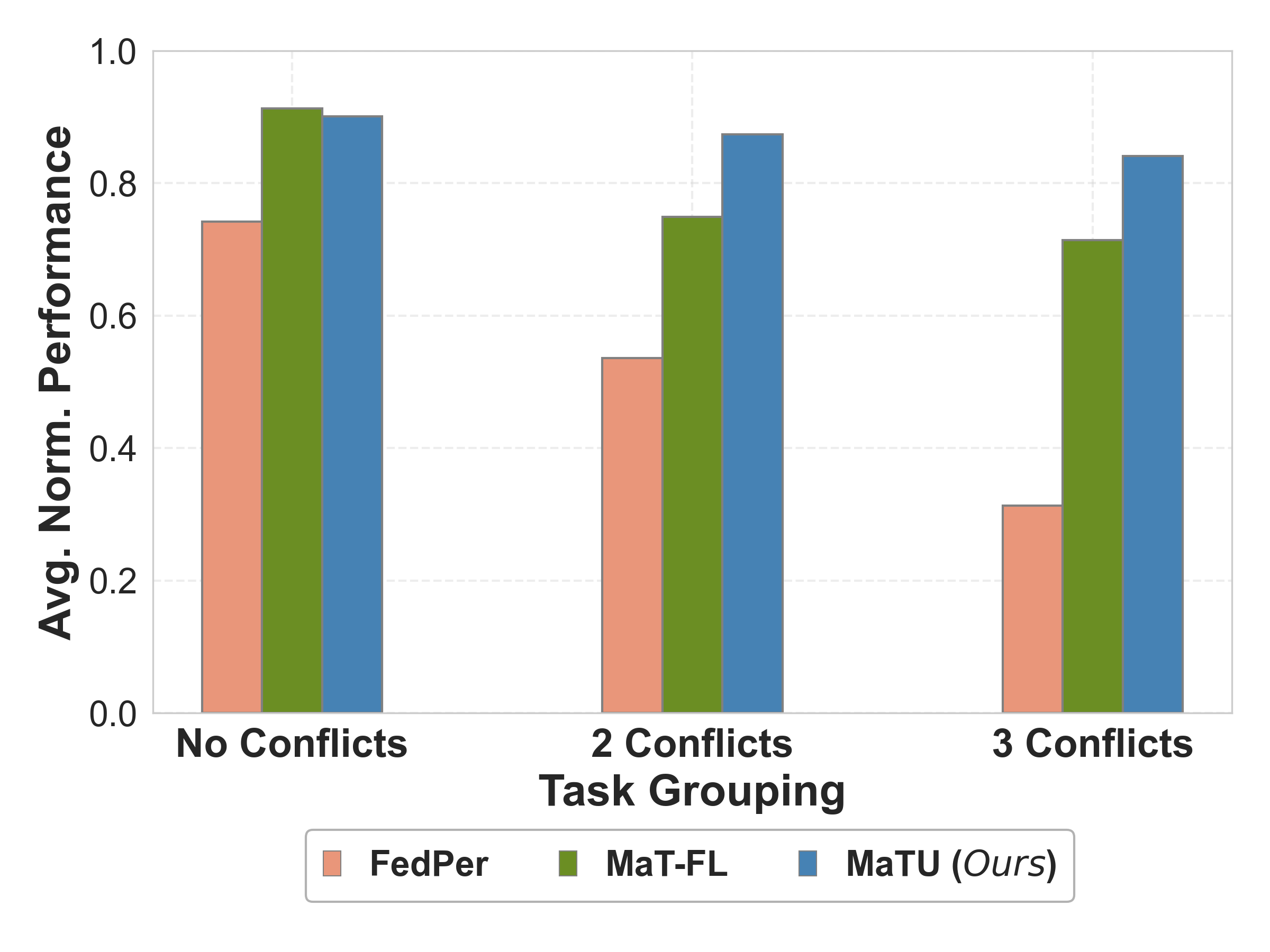}
            \vspace{-15pt}
            \caption{Task conflict groups\label{fig:conflict_tasks}}
        \end{subfigure}
    \end{minipage}\hfill
    \begin{minipage}{0.495\linewidth}
        \begin{subfigure}[b]{1.0\linewidth}
            \centering
            \vspace{6pt}            
            \includegraphics[width=\textwidth]{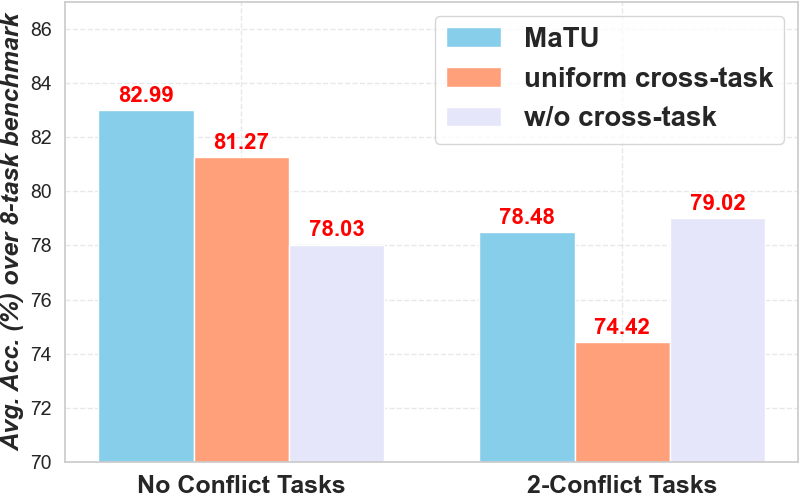}
            \vspace{-4pt}
            \caption{Cross-task aggregation\label{fig:cross_task}}
        \end{subfigure}
    \end{minipage}
    \vspace{-4pt}
    \caption{Performance evaluation of ViT-B/32 on the $8$-task benchmark under (a) task conflict groups and (b) cross-task aggregation variants. We report (a) normalized test set accuracy vs individual task fine-tuning and (b) average accuracy on test set. Federated parameters are set to $N$=$10$, $R$=$300$, $\xi$=$1$ and $E$=$1$.}
    \vspace{-10pt}
\end{figure}

\vspace{-5pt}
\noindent \textit{$8$-task benchmark}: Table~\ref{tab:multitask_perf} demonstrates that~\method~maintains strong performance in \textit{multiple-task} settings, achieving an average accuracy of 79.47\%~\textendash~a modest 5\% drop from single-task settings~\textendash~while outperforming all baselines and effectively minimizing task interference via masking. Traditional FL methods, such as FedAvg and FedProx, exhibit significant performance deficits, with average accuracies well below their single-task counterparts, underscoring their limitations in addressing \textit{task heterogeneity}. Notably, NTK-FedAvg surpasses these baselines, likely benefiting from improved task disentanglement through NTK-based linearization. In stark contrast, FedPer experiences a severe performance decline ($\approx$40\%) when transitioning from \textit{single} to \textit{multiple-task} settings, reflecting its inability to manage intra-client \textit{task heterogeneity} due to its reliance on personalization layers. This highlights that personalized FL approaches are not designed to address the broader challenges of \textit{task heterogeneity}, underscoring that personalization represents a subset of the more general MaT-FL problem. Conversely, \textit{MaT-FL} achieves higher performance by employing similarity-based grouping to mitigate task conflicts, reducing interactions among dissimilar clients and tasks, but still lags~\method~by 10\%. We further analyze how \textit{MaT-FL} and~\method~handle such conflicts in Fig.˜\ref{fig:conflict_tasks}. 


Beyond test accuracy,~\method~offers substantial communication and storage efficiency in \textit{multiple-task} settings. As shown by the $\mathrm{bpt}$ column in Table~\ref{tab:multitask_perf}, competing methods transmit multiple adapters — one per task—leading to high communication costs and server-side storage of client- or group-specific weights. In contrast,~\method~transmits only a single unified task vector per client, along with lightweight modulators (a binary mask and a scalar per task), enabling stateless aggregation and scalable deployment without storing any client-specific parameters. \\

\begin{figure}[!t]
    \centering \small
    \begin{minipage}{0.49\linewidth}
        \begin{subfigure}[b]{1.0\linewidth}
            \centering
            \includegraphics[width=\textwidth]{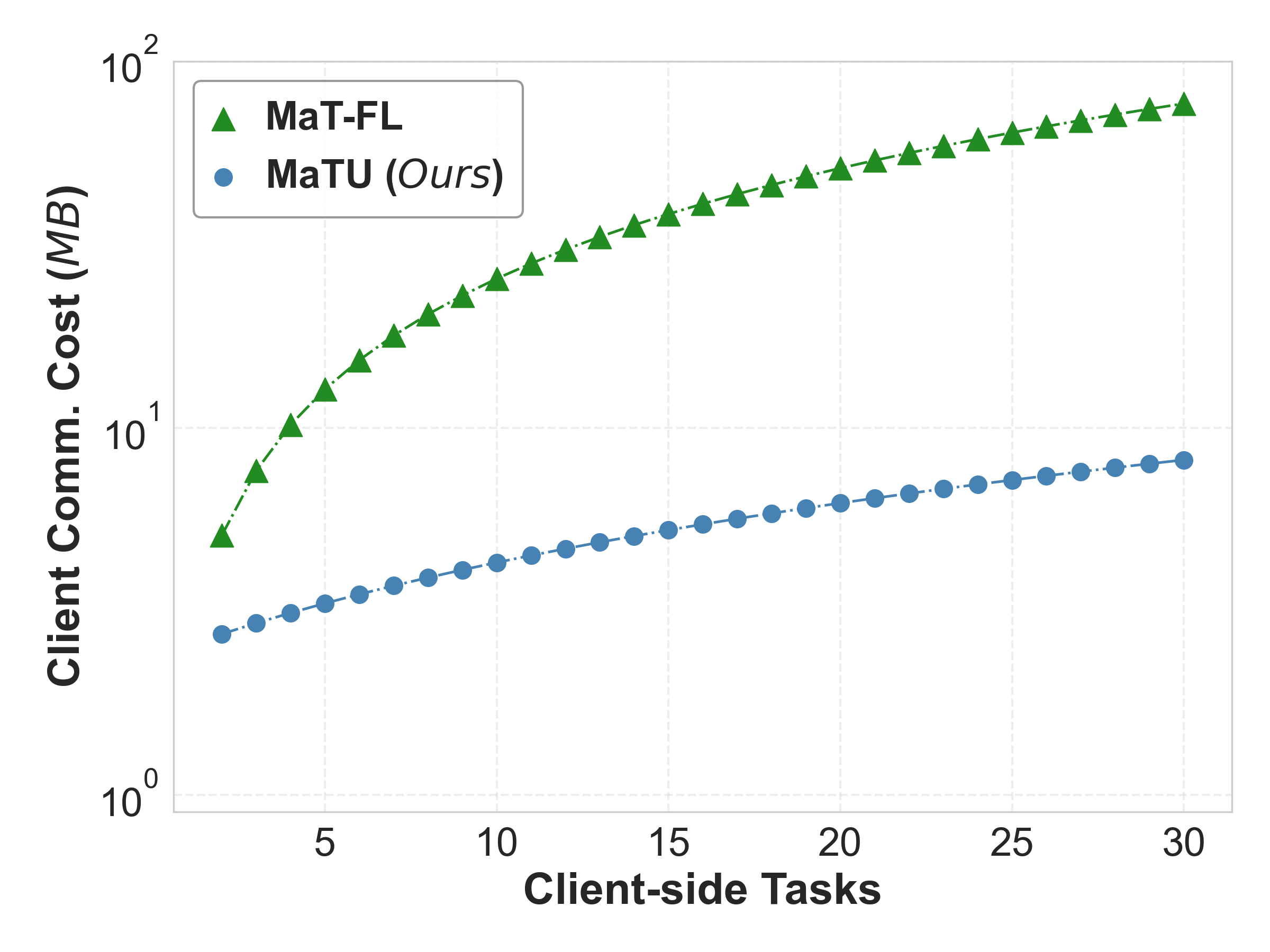}
            \caption{Comm. Overhead\label{fig:comm_savings}}
        \end{subfigure}
    \end{minipage}\hfill
    \begin{minipage}{0.49\linewidth}
        \begin{subfigure}[b]{1.0\linewidth}
            \centering
            \includegraphics[width=\textwidth]{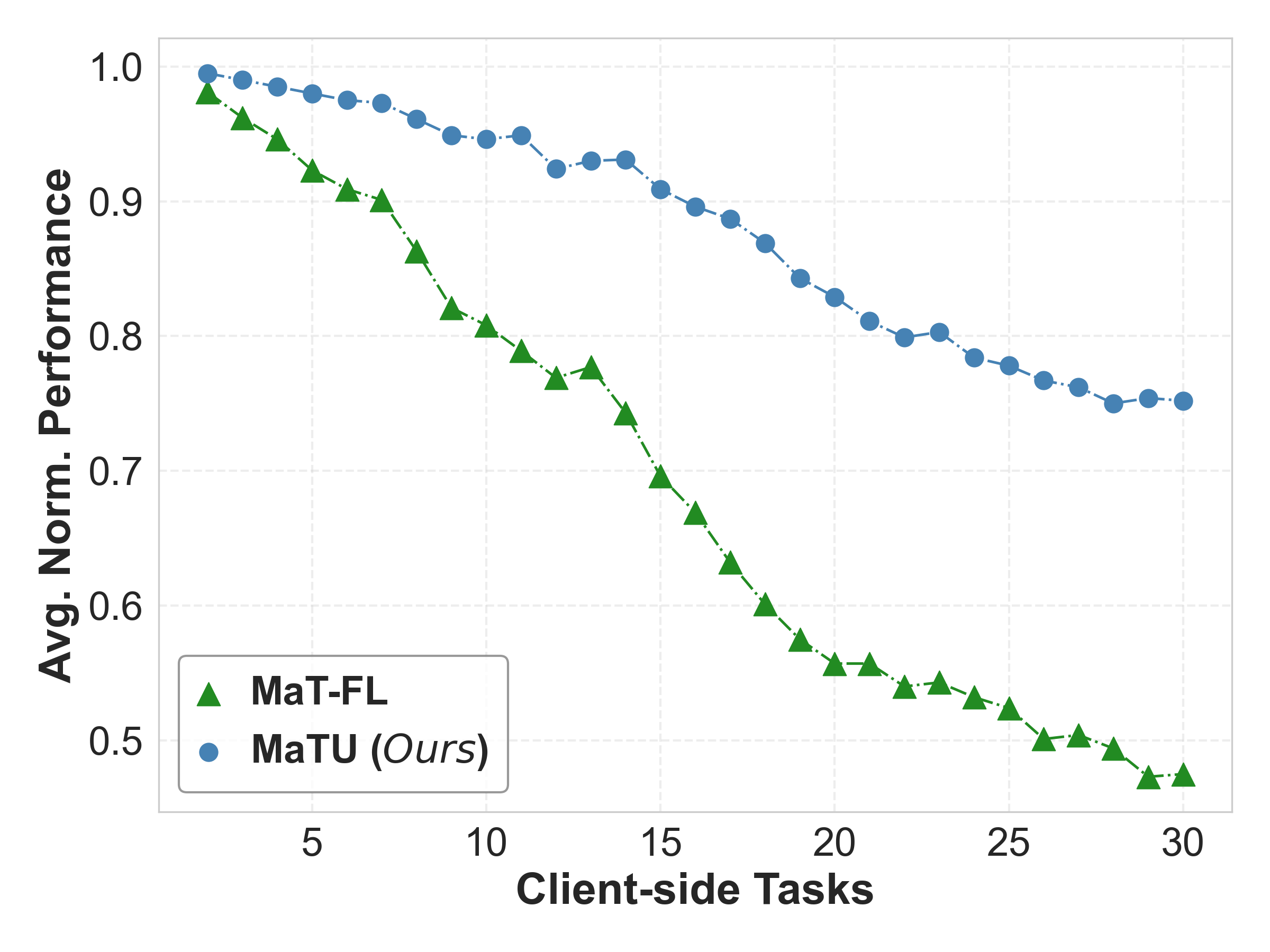}
            \caption{Norm. Performance.\label{fig:norm_perf}}
        \end{subfigure}
    \end{minipage}
    \vspace{-4pt}
    \caption{Impact of scaling number of tasks assigned to clients on~\method. Performance evaluation of ViT-B/32 on the $30$-task benchmark. We report (a) communication cost for 1 federated round, and (b) normalized test set accuracy vs individual task fine-tuning. Federated parameters are set to $N$=$10$, $R$=$300$, $\xi$=$1$ and $E$=$1$.\label{fig:task_scaling}}
    \vspace{-10pt}
\end{figure}

    
\vspace{-5pt}
\noindent \textit{$30$-task benchmark}: We evaluate~\method’s scalability to $30$ tasks and compare it to \textit{MaT-FL}, the strongest-performing baseline in \textit{multiple-task} settings (see Table~\ref{tab:multitask_perf}). As shown in Fig.\ref{fig:30_vit_tasks},~\method~achieves an average normalized performance of 77.40\%, significantly surpassing \textit{MaT-FL}’s 52.62\% across most datasets, except MangoLeafDB. This demonstrates~\method’s ability to minimize task interference through training and delivers performance comparable to individual task fine-tuning. An elaborate analysis of~\method’s effectiveness with respect of increase of client's tasks is presented in Fig.~\ref{fig:task_scaling}. More importantly, beyond superior joint training performance,~\method~constructs a ``\textit{unified}" task vector that encapsulates all tasks and can be swiftly adapted to individual tasks using computationally lightweight \textit{modulators}, enabling significant storage savings during inference, particularly as model size scale. \\



\noindent \textbf{Scaling clients tasks.} We further analyze how~\method~handles an increasing number of tasks per client. Specifically, we conduct experiments on the $30$-task benchmark, varying each client’s task groups from $2$ to $30$, and report both (i) communication cost per round (in MB) and (ii) average normalized performance across tasks (vs individual task fine-tuning). As shown in Fig.\ref{fig:comm_savings},~\method~introduces minimal communication overhead as client tasks scale. This efficiency arises from transmitting a single adapter per client, regardless of number of tasks, supplemented by lightweight task-specific modulators (a binary mask and a scalar value), unlike the multiple adapters required by\textit{MaT-FL} and other baselines. Furthermore, in terms of model performance, Fig.\ref{fig:norm_perf} shows that~\textit{MaT-FL}~suffers from a sharp performance drop when clients are assigned more than $5$ tasks, highlighting its limitations in managing high \textit{task heterogeneity}. In contrary,~\method~demonstrates to have a solid performance~\textendash~even when clients train on over $15$ tasks~\textendash~highlighting its ability to effectively disentangle tasks in the weight space through task-specific modulators.\\

\noindent \textbf{Highly Conflicting task groups.} Here, we evaluate~\method’s performance in scenarios with highly conflicting tasks within client's task group. Using the $8$-task benchmark, we focus on the $3$ distinct task clusters identified in Fig.\ref{fig:sim_matrices}. Specifically, we conduct experiments with $10$ clients, each assigned a fixed group of three tasks, designed to include no, $2$, or $3$ highly dissimilar tasks, referred to as \textit{no conflict}, \textit{2-conflict}, and \textit{3-conflict} task groups, respectively. This setup enables us to assess \textit{how~\method~handles task conflicts on the clients} and compare its performance against baselines. As shown in Fig.~\ref{fig:conflict_tasks},~\method~achieves consistently high performance with minimal drop (less than $6$\%), unlike baselines such as FedPer, which struggles to train multi-task models even without conflicting task groups, and \textit{MaT-FL}, which suffers from a more pronounced performance drop as task conflicts increase. \\

\noindent \textbf{Cross-task aggregation.} We also analyze the benefits of cross-task aggregation (Eq.~6) by comparing~\method~against two variants: one without cross-task aggregation and another using uniform cross-task averaging, where all task vectors are treated equally regardless of similarity. As shown in Fig.\ref{fig:cross_task},~\method~achieves superior performance in the absence of conflicts and effectively mitigates negative transfer under \textit{2-conflict} settings. In contrast, the uniform variant degrades under task heterogeneity, while removing cross-task aggregation entirely limits performance.

\section{Conclusions}
In this work, we introduced~\method, a MaT-FL approach designed to address task heterogeneity by relying on task vectors that creates a ``\textit{unified}'' task vector~\textendash~encapsulating all tasks~\textendash~together with lightweight task modulators. By eliminating the need for clustering or client-specific weight storage,~\method~ enables scalable, low-overhead MaT-FL even with large numbers of tasks. Our evaluations across $30$ datasets demonstrated that~\method~not only achieves superior performance compared to state-of-the-art MaT-FL regimes but also delivers results comparable to per-task fine-tuning while significantly reducing communication overhead; thus highlighting~\method's potential under highly heterogeneous FL cases.

\vspace{-5pt}
\small{
\section*{Acknowledgments}
This work was funded by the DAIS project, which has received funding from ECSEL Joint Undertaking under grant agreement No 101007273, and the Bio-Curity project, which has received funding from Eureka Cluster Xecs under grant agreement No 2022016.
}

\clearpage
\bibliographystyle{named}
\bibliography{refs}

\begin{thebibliography}{}

\bibitem[\protect\citeauthoryear{Amari}{1998}]{Amari1998NaturalGW}
Shun‐ichi Amari.
\newblock Natural gradient works efficiently in learning.
\newblock {\em Neural Computation}, 10:251--276, 1998.

\bibitem[\protect\citeauthoryear{Arivazhagan \bgroup \em et al.\egroup }{2019}]{arivazhagan2019federatedlearningpersonalizationlayers}
Manoj~Ghuhan Arivazhagan, Vinay Aggarwal, Aaditya~Kumar Singh, and Sunav Choudhary.
\newblock Federated learning with personalization layers, 2019.

\bibitem[\protect\citeauthoryear{Babakniya \bgroup \em et al.\egroup }{2023}]{babakniya2023slorafederatedparameterefficient}
Sara Babakniya, Ahmed~Roushdy Elkordy, Yahya~H. Ezzeldin, Qingfeng Liu, Kee-Bong Song, Mostafa El-Khamy, and Salman Avestimehr.
\newblock Slora: Federated parameter efficient fine-tuning of language models, 2023.

\bibitem[\protect\citeauthoryear{Bao \bgroup \em et al.\egroup }{2022}]{bao2022informationtheoreticapproachtransferabilitytask}
Yajie Bao, Yang Li, Shao-Lun Huang, Lin Zhang, Lizhong Zheng, Amir Zamir, and Leonidas Guibas.
\newblock An information-theoretic approach to transferability in task transfer learning, 2022.

\bibitem[\protect\citeauthoryear{Beutel \bgroup \em et al.\egroup }{2020}]{beutel2020flower}
Daniel~J Beutel, Taner Topal, Akhil Mathur, Xinchi Qiu, Javier Fernandez-Marques, Yan Gao, Lorenzo Sani, Hei~Li Kwing, Titouan Parcollet, Pedro PB~de Gusmão, and Nicholas~D Lane.
\newblock Flower: A friendly federated learning research framework.
\newblock {\em arXiv preprint arXiv:2007.14390}, 2020.

\bibitem[\protect\citeauthoryear{Cai \bgroup \em et al.\egroup }{2023}]{10208649}
Ruisi Cai, Xiaohan Chen, Shiwei Liu, Jayanth Srinivasa, Myungjin Lee, Ramana Kompella, and Zhangyang Wang.
\newblock Many-task federated learning: A new problem setting and a simple baseline.
\newblock In {\em 2023 IEEE/CVF Conference on Computer Vision and Pattern Recognition Workshops (CVPRW)}, pages 5037--5045, 2023.

\bibitem[\protect\citeauthoryear{Chen \bgroup \em et al.\egroup }{2023}]{chen2023fedbone}
Yiqiang Chen, Teng Zhang, Xinlong Jiang, Qian Chen, Chenlong Gao, and Wuliang Huang.
\newblock Fedbone: Towards large-scale federated multi-task learning.
\newblock {\em arXiv preprint arXiv:2306.17465}, 2023.

\bibitem[\protect\citeauthoryear{Cheng \bgroup \em et al.\egroup }{2017}]{resisc45}
Gong Cheng, Junwei Han, and Xiaoqiang Lu.
\newblock Remote sensing image scene classification: Benchmark and state of the art.
\newblock {\em Proceedings of the IEEE}, 105(10):1865--1883, 2017.

\bibitem[\protect\citeauthoryear{Cho \bgroup \em et al.\egroup }{2023}]{cho2024heterogeneous}
Yae~Jee Cho, Luyang Liu, Zheng Xu, Aldi Fahrezi, Matt Barnes, and Gauri Joshi.
\newblock Heterogeneous lo{RA} for federated fine-tuning of on-device foundation models.
\newblock In {\em International Workshop on Federated Learning in the Age of Foundation Models in Conjunction with NeurIPS 2023}, 2023.

\bibitem[\protect\citeauthoryear{Cimpoi \bgroup \em et al.\egroup }{2014}]{dtd}
M.~Cimpoi, S.~Maji, I.~Kokkinos, S.~Mohamed, and A.~Vedaldi.
\newblock Describing textures in the wild.
\newblock In {\em Proceedings of the {IEEE} Conf. on Computer Vision and Pattern Recognition ({CVPR})}, 2014.

\bibitem[\protect\citeauthoryear{Deng \bgroup \em et al.\egroup }{2009}]{5206848}
Jia Deng, Wei Dong, Richard Socher, Li-Jia Li, Kai Li, and Li~Fei-Fei.
\newblock Imagenet: A large-scale hierarchical image database.
\newblock In {\em 2009 IEEE Conference on Computer Vision and Pattern Recognition}, pages 248--255, 2009.

\bibitem[\protect\citeauthoryear{Ding \bgroup \em et al.\egroup }{2022}]{ding2022pactranpacbayesianmetricsestimating}
Nan Ding, Xi~Chen, Tomer Levinboim, Beer Changpinyo, and Radu Soricut.
\newblock Pactran: Pac-bayesian metrics for estimating the transferability of pretrained models to classification tasks, 2022.

\bibitem[\protect\citeauthoryear{Dosovitskiy \bgroup \em et al.\egroup }{2021}]{dosovitskiy2021imageworth16x16words}
Alexey Dosovitskiy, Lucas Beyer, Alexander Kolesnikov, Dirk Weissenborn, Xiaohua Zhai, Thomas Unterthiner, Mostafa Dehghani, Matthias Minderer, Georg Heigold, Sylvain Gelly, Jakob Uszkoreit, and Neil Houlsby.
\newblock An image is worth 16x16 words: Transformers for image recognition at scale, 2021.

\bibitem[\protect\citeauthoryear{Gholami \bgroup \em et al.\egroup }{2023}]{gholami2023etranenergybasedtransferabilityestimation}
Mohsen Gholami, Mohammad Akbari, Xinglu Wang, Behnam Kamranian, and Yong Zhang.
\newblock Etran: Energy-based transferability estimation, 2023.

\bibitem[\protect\citeauthoryear{Helber \bgroup \em et al.\egroup }{2017}]{eurosat}
Patrick Helber, Benjamin Bischke, Andreas Dengel, and Damian Borth.
\newblock Eurosat: A novel dataset and deep learning benchmark for land use and land cover classification, 2017.

\bibitem[\protect\citeauthoryear{Hu \bgroup \em et al.\egroup }{2021}]{hu2021loralowrankadaptationlarge}
Edward~J. Hu, Yelong Shen, Phillip Wallis, Zeyuan Allen-Zhu, Yuanzhi Li, Shean Wang, Lu~Wang, and Weizhu Chen.
\newblock Lora: Low-rank adaptation of large language models, 2021.

\bibitem[\protect\citeauthoryear{Huang \bgroup \em et al.\egroup }{2024}]{huang2024emrmergingtuningfreehighperformancemodel}
Chenyu Huang, Peng Ye, Tao Chen, Tong He, Xiangyu Yue, and Wanli Ouyang.
\newblock Emr-merging: Tuning-free high-performance model merging, 2024.

\bibitem[\protect\citeauthoryear{Ilharco \bgroup \em et al.\egroup }{2023}]{ilharco2023editing}
Gabriel Ilharco, Marco~Tulio Ribeiro, Mitchell Wortsman, Suchin Gururangan, Ludwig Schmidt, Hannaneh Hajishirzi, and Ali Farhadi.
\newblock Editing models with task arithmetic, 2023.

\bibitem[\protect\citeauthoryear{Krause \bgroup \em et al.\egroup }{2013}]{cars}
Jonathan Krause, Michael Stark, Jia Deng, and Li~Fei-Fei.
\newblock 3d object representations for fine-grained categorization.
\newblock In {\em 2013 IEEE International Conference on Computer Vision Workshops}, pages 554--561, 2013.

\bibitem[\protect\citeauthoryear{Kuang \bgroup \em et al.\egroup }{2023}]{kuang2023federatedscopellmcomprehensivepackagefinetuning}
Weirui Kuang, Bingchen Qian, Zitao Li, Daoyuan Chen, Dawei Gao, Xuchen Pan, Yuexiang Xie, Yaliang Li, Bolin Ding, and Jingren Zhou.
\newblock Federatedscope-llm: A comprehensive package for fine-tuning large language models in federated learning, 2023.

\bibitem[\protect\citeauthoryear{LeCun \bgroup \em et al.\egroup }{2010}]{mnist}
Yann LeCun, Corinna Cortes, and CJ~Burges.
\newblock Mnist handwritten digit database.
\newblock {\em ATT Labs [Online]. Available: http://yann.lecun.com/exdb/mnist}, 2, 2010.

\bibitem[\protect\citeauthoryear{Li \bgroup \em et al.\egroup }{2020}]{li2020federatedoptimizationheterogeneousnetworks}
Tian Li, Anit~Kumar Sahu, Manzil Zaheer, Maziar Sanjabi, Ameet Talwalkar, and Virginia Smith.
\newblock Federated optimization in heterogeneous networks, 2020.

\bibitem[\protect\citeauthoryear{Li \bgroup \em et al.\egroup }{2021}]{li2021federatedlearningnoniiddata}
Qinbin Li, Yiqun Diao, Quan Chen, and Bingsheng He.
\newblock Federated learning on non-iid data silos: An experimental study, 2021.

\bibitem[\protect\citeauthoryear{Liu \bgroup \em et al.\egroup }{2025}]{LIU2025106796}
Xinran Liu, Yikun Bai, Yuzhe Lu, Andrea Soltoggio, and Soheil Kolouri.
\newblock Wasserstein task embedding for measuring task similarities.
\newblock {\em Neural Networks}, 181:106796, 2025.

\bibitem[\protect\citeauthoryear{Lu \bgroup \em et al.\egroup }{2023}]{lu2023towards}
Yuxiang Lu, Suizhi Huang, Yuwen Yang, Shalayiding Sirejiding, Yue Ding, and Hongtao Lu.
\newblock Towards hetero-client federated multi-task learning.
\newblock {\em arXiv preprint arXiv:2311.13250}, 2023.

\bibitem[\protect\citeauthoryear{McMahan \bgroup \em et al.\egroup }{2017}]{mcmahan2017communication}
Brendan McMahan, Eider Moore, Daniel Ramage, Seth Hampson, and Blaise~Aguera y~Arcas.
\newblock Communication-efficient learning of deep networks from decentralized data.
\newblock In {\em Artificial intelligence and statistics}, pages 1273--1282. PMLR, 2017.

\bibitem[\protect\citeauthoryear{Muhamed \bgroup \em et al.\egroup }{2024}]{muhamed2024fed}
Aashiq Muhamed, Meher Mankikar, and Virginia Smith.
\newblock Fed up with complexity: Simplifying many-task federated learning with {NTKF}edavg.
\newblock In {\em Privacy Regulation and Protection in Machine Learning}, 2024.

\bibitem[\protect\citeauthoryear{Netzer \bgroup \em et al.\egroup }{2011}]{svhn}
Yuval Netzer, Tao Wang, Adam Coates, Alessandro Bissacco, Bo~Wu, and Andrew~Y Ng.
\newblock Reading digits in natural images with unsupervised feature learning.
\newblock 2011.

\bibitem[\protect\citeauthoryear{Nguyen \bgroup \em et al.\egroup }{2024}]{nguyen2024flora}
Duy~Phuong Nguyen, J~Pablo Munoz, and Ali Jannesari.
\newblock Flora: Enhancing vision-language models with parameter-efficient federated learning.
\newblock {\em arXiv preprint arXiv:2404.15182}, 2024.

\bibitem[\protect\citeauthoryear{Ortiz-Jimenez \bgroup \em et al.\egroup }{2023}]{ortizjimenez2023task}
Guillermo Ortiz-Jimenez, Alessandro Favero, and Pascal Frossard.
\newblock Task arithmetic in the tangent space: Improved editing of pre-trained models, 2023.

\bibitem[\protect\citeauthoryear{Ping \bgroup \em et al.\egroup }{2024}]{ping2024fl}
Siqi Ping, Yuzhu Mao, Yang Liu, Xiao-Ping Zhang, and Wenbo Ding.
\newblock Fl-tac: Enhanced fine-tuning in federated learning via low-rank, task-specific adapter clustering.
\newblock {\em arXiv preprint arXiv:2404.15384}, 2024.

\bibitem[\protect\citeauthoryear{Stallkamp \bgroup \em et al.\egroup }{2011}]{gtsrb}
Johannes Stallkamp, Marc Schlipsing, Jan Salmen, and Christian Igel.
\newblock The german traffic sign recognition benchmark: A multi-class classification competition.
\newblock In {\em The 2011 International Joint Conference on Neural Networks}, pages 1453--1460, 2011.

\bibitem[\protect\citeauthoryear{Sun \bgroup \em et al.\egroup }{2024}]{sun2024improvingloraprivacypreservingfederated}
Youbang Sun, Zitao Li, Yaliang Li, and Bolin Ding.
\newblock Improving lora in privacy-preserving federated learning, 2024.

\bibitem[\protect\citeauthoryear{Tenison \bgroup \em et al.\egroup }{2023}]{tenison2023gradientmaskedaveragingfederated}
Irene Tenison, Sai~Aravind Sreeramadas, Vaikkunth Mugunthan, Edouard Oyallon, Irina Rish, and Eugene Belilovsky.
\newblock Gradient masked averaging for federated learning, 2023.

\bibitem[\protect\citeauthoryear{Tsouvalas \bgroup \em et al.\egroup }{2023}]{tsouvalas2023federatedfinetuningfoundationmodels}
Vasileios Tsouvalas, Yuki Asano, and Aaqib Saeed.
\newblock Federated fine-tuning of foundation models via probabilistic masking, 2023.

\bibitem[\protect\citeauthoryear{Vu \bgroup \em et al.\egroup }{2022}]{vu2022spotbetterfrozenmodel}
Tu~Vu, Brian Lester, Noah Constant, Rami Al-Rfou, and Daniel Cer.
\newblock Spot: Better frozen model adaptation through soft prompt transfer, 2022.

\bibitem[\protect\citeauthoryear{{Xiao} \bgroup \em et al.\egroup }{2010}]{sun397}
J.~{Xiao}, J.~{Hays}, K.~A. {Ehinger}, A.~{Oliva}, and A.~{Torralba}.
\newblock Sun database: Large-scale scene recognition from abbey to zoo.
\newblock In {\em 2010 IEEE Computer Society Conference on Computer Vision and Pattern Recognition}, pages 3485--3492, June 2010.

\bibitem[\protect\citeauthoryear{Yadav \bgroup \em et al.\egroup }{2023a}]{yadav2023tiesmerging}
Prateek Yadav, Derek Tam, Leshem Choshen, Colin Raffel, and Mohit Bansal.
\newblock Ties-merging: Resolving interference when merging models, 2023.

\bibitem[\protect\citeauthoryear{Yadav \bgroup \em et al.\egroup }{2023b}]{yadav2023tiesmergingresolvinginterferencemerging}
Prateek Yadav, Derek Tam, Leshem Choshen, Colin Raffel, and Mohit Bansal.
\newblock Ties-merging: Resolving interference when merging models, 2023.

\bibitem[\protect\citeauthoryear{Yang \bgroup \em et al.\egroup }{2024}]{yang2024adamergingadaptivemodelmerging}
Enneng Yang, Zhenyi Wang, Li~Shen, Shiwei Liu, Guibing Guo, Xingwei Wang, and Dacheng Tao.
\newblock Adamerging: Adaptive model merging for multi-task learning, 2024.

\bibitem[\protect\citeauthoryear{Yi \bgroup \em et al.\egroup }{2023}]{yi2023fedlora}
Liping Yi, Han Yu, Gang Wang, and Xiaoguang Liu.
\newblock Fedlora: Model-heterogeneous personalized federated learning with lora tuning.
\newblock {\em arXiv preprint arXiv:2310.13283}, 2023.

\bibitem[\protect\citeauthoryear{Yu \bgroup \em et al.\egroup }{2024}]{yu2024languagemodelssupermario}
Le~Yu, Bowen Yu, Haiyang Yu, Fei Huang, and Yongbin Li.
\newblock Language models are super mario: Absorbing abilities from homologous models as a free lunch, 2024.

\bibitem[\protect\citeauthoryear{Zamir \bgroup \em et al.\egroup }{2018}]{zamir2018taskonomydisentanglingtasktransfer}
Amir Zamir, Alexander Sax, William Shen, Leonidas Guibas, Jitendra Malik, and Silvio Savarese.
\newblock Taskonomy: Disentangling task transfer learning, 2018.

\bibitem[\protect\citeauthoryear{Zhuang \bgroup \em et al.\egroup }{2023}]{zhuang2023mas}
Weiming Zhuang, Yonggang Wen, Lingjuan Lyu, and Shuai Zhang.
\newblock Mas: Towards resource-efficient federated multiple-task learning.
\newblock In {\em Proceedings of the IEEE/CVF International Conference on Computer Vision}, pages 23414--23424, 2023.

\end{thebibliography}

\end{document}